\documentclass[draft]{agujournal2019}

\usepackage{amsmath}
\usepackage{graphicx}
\usepackage{float}
\usepackage[finalnew]{trackchanges}
\usepackage[T1]{fontenc}
\usepackage[utf8]{inputenc}
\usepackage{tabularx}
\usepackage{booktabs}
\usepackage{multirow}
\usepackage{multicol}
\usepackage{indentfirst}
\usepackage{xcolor}
\usepackage[a4paper]{geometry}
\usepackage{lmodern}
\usepackage{makecell}
\newcommand{\modified}{\textcolor{black}}

\newcommand{\changed}{\textcolor{black}}
\newcommand{\newchanged}{\textcolor{black}}

\usepackage{url} 
\usepackage{lineno}
\usepackage{soul}
\usepackage{algorithm}

\draftfalse

\journalname{JGR: Machine Learning and Computation}

\begin{document}

%
%
\title{Comparative and Interpretative Analysis of CNN and Transformer Models in Predicting Wildfire Spread Using Remote Sensing Data}

%
%




\authors{\authors{Yihang Zhou\affil{1}, Ruige Kong\affil{2},Zhengsen Xu\affil{3}, Linlin Xu\affil{4} and Sibo Cheng*\affil{5}}}

\affiliation{1}{Department of Earth Science and Engineering, Imperial College London, London, UK}
\affiliation{2}{Department of Engineering, University of Cambridge, Cambridge, UK}
\affiliation{3}{Department of Geography and Environmental Management, University of Waterloo, ON, Canada}
\affiliation{4}{Systems Design Engineering, University of Waterloo, Waterloo, ON, Canada}
\affiliation{5}{CEREA, ENPC and EDF R\&D, Institut Polytechnique de Paris, \^Ile-de-France, France}




\correspondingauthor{Sibo Cheng}{sibo.cheng@enpc.fr}



\begin{keypoints}
\item Our comprehensive analysis compared four prevalent deep learning models for predicting wildfires using a decade-long dataset from the U.S.
\item Our comparative analysis demonstrates that UNet and Transformer-based Swin-UNet outperform conventional CNN methods in wildfire prediction. 
\item We enhanced model transparency and understood their characteristics, providing insights for optimal model selection and future improvements.
\end{keypoints}

%
%

%
%


\begin{abstract}

Facing the escalating threat of global wildfires, numerous computer vision techniques using remote sensing data have been applied in this area. However, the selection of deep learning methods for wildfire prediction remains uncertain due to the lack of comparative analysis in a quantitative and explainable manner, crucial for improving prevention measures and refining models. This study aims to thoroughly compare the performance, efficiency, and explainability of four prevalent deep learning architectures: Autoencoder, ResNet, UNet, and Transformer-based Swin-UNet. Employing a real-world dataset that includes nearly a decade of remote sensing data from California, U.S., these models predict the spread of wildfires for the following day. Through detailed quantitative comparison analysis, we discovered that Transformer-based Swin-UNet and UNet generally outperform Autoencoder and ResNet, particularly due to the advanced attention mechanisms in Transformer-based Swin-UNet and the efficient use of skip connections in both UNet and Transformer-based Swin-UNet, which contribute to superior predictive accuracy and model interpretability. Then we applied XAI
techniques on all four models, this not only enhances the clarity and trustworthiness of models but also promotes focused improvements in wildfire prediction capabilities. The XAI analysis reveals that UNet and Transformer-based Swin-UNet are able to focus on critical features such as 'Previous Fire Mask', 'Drought', and 'Vegetation' more effectively than the other two models, while also maintaining balanced attention to the remaining features, leading to their superior performance. The insights from our thorough comparative analysis offer substantial implications for future model design and also provide guidance for model selection in different scenarios. The source code for this project is publicly available as open source at \url{https://github.com/Yng314/xai_for_wildfire.}
\end{abstract}

\section*{Plain Language Summary}
As wildfires increase globally, predicting their occurrence accurately and understandably has become more critical than ever. This study compared advanced computer models using a decade of space-based data to predict next-day wildfire risks in the United States. We focused on two types of models: Convolutional Neural Networks and Transformer models. A key aspect of our research was not only determining which model predicts wildfires more accurately but also understanding how these models make their predictions. This is where XAI plays a crucial role. XAI helps us explore these complex models to see how they process and interpret data, ensuring that the predictions they make are both reliable and transparent. Through detailed comparisons and analyses, our findings highlight significant differences in model performance and their approaches to interpreting data. Emphasizing both accuracy and explainability, this study enhances our ability to select and refine the best models for predicting wildfires, offering crucial insights that could improve prevention strategies and advance wildfire prediction technology.

\section{Introduction}
\subsection{Background}
Wildfires pose a significant threat to ecosystems, property, and human life worldwide \cite{nolan2022increasing, bousfield2023substantial}. Accurate and timely prediction of these natural disasters is essential for effective prevention and management strategies. In recent years, the emergence of remote sensing technology has revolutionized wildfire monitoring and prediction. Remote sensing data, including a range of spectral, spatial, and temporal resolutions, provide critical information on vegetation status, moisture content, topography, and other environmental factors instrumental in wildfire development and spread.

The integration of deep learning models with remote sensing data has emerged as a promising approach to wildfire prediction and analysis. Deep learning models, such as autoencoders \cite{huot2021next}, ResNet \cite{he2016deep}, UNet \cite{ronneberger2015u}, and Vision Transformers (ViT) \cite{dosovitskiy2020image}, have demonstrated substantial capabilities in processing complex spatial and temporal data, offering significant advantages in efficiency and accuracy over traditional statistical methods. These models can identify subtle patterns and correlations in large datasets, \changed{facilitating more accurate predictions of wildfire occurrences \cite{Suwansrikham2023Performance, Khryashchev2020Wildfire} and behavior \cite{Ivek2022Reconstruction}}. \remove{For example, Al-Dabbaghand Ilyas (2023) processed uni-temporal Sentinel-2 imagery using UNet with ResNet50, achieving an F1-score of 98.78\% and an IoU of 97.38\% in the semantic segmentation of wildfire-affected areas. The dataset Al-Dabbaghand Ilyas (2023) used was captured by the Multi Spectral Instrument sensor and has a spatial resolution of 10 to 60 meters, which highlights its spatial complexity.} \changed{For example, \citeA{Al-Dabbagh2023Uni-temporal} processed uni-temporal Sentinel-2 imagery using UNet with ResNet50, achieving an F1-score of 98.78\% and an IoU of 97.38\% in the semantic segmentation of wildfire-affected areas. The dataset used in this study, captured by the Multi Spectral Instrument sensor, has a spatial resolution of 10 meters and consists of data from five non-continuous days in September 2021, highlighting its temporal and spatial complexity.} \remove{Similarly, Suwansrikham and Singkhamfu (2023) utilized the ViT model to process UAV aerial photography data, achieving the highest accuracy of 88.03\% in forest fire detection. This dataset, consisting of 4,450 images, demonstrates the capability of the Transformer-based model to handle and interpret complex spatial data.} \changed{Similarly, \citeA{Suwansrikham2023Performance} utilized the ViT model to process UAV aerial photography data, achieving the highest accuracy of 88.03\% in forest fire detection. This dataset consists of 56 historical fire georeferenced perimeters from the period of 2014–2016, with a spatial resolution of 30 meters, demonstrating the capability of the \\Transformer-based model to handle and interpret complex spatial data effectively.}

However, the complexity of deep learning models often leads to a 'black box' problem where the decision-making process is neither transparent nor understandable \cite{castelvecchi2016can}. \modified{This lack of explainability can be a significant barrier, especially in high-stakes scenarios such as wildfire prediction, where understanding the rationale behind predictions is crucial for trustworthy and actionable insights and for guiding subsequent modeling and decision-making.} \cite{Kondylatos2022Wildfire, Girtsou2021A}. For example, \citeA{ABDOLLAHI2023163004} underscored the need for explainability in wildfire prediction models as their outputs guide natural resource management plans, necessitating an understanding of the underlying logic by decision-makers. \citeA{fan2024explainable} also emphasized the importance of explainability when facilitating more effective and informed forest fire management strategies. Therefore, in this paper, to address the critical role of model explainability, we \remove{creatively} employ a suite of interpretability methods: SHapley Additive exPlanations (SHAP) \cite{lundberg2017unified}, Gradient-weighted Class Activation Mapping (Grad-CAM) \cite{sundararajan2017axiomatic}, and Integrated Gradients (IG) \cite{selvaraju2017grad}, to dissect and illuminate the decision-making process of our predictive models. \modified{Beyond merely applying these tools, we conduct a detailed analysis to uncover the most influential features for wildfire propagation, and how these models prioritize the features in the dataset, helping us explore their characteristics and reveal their strengths and weaknesses.}

\begin{figure*}[!ht]
    \centering
    \includegraphics[width=1\textwidth]{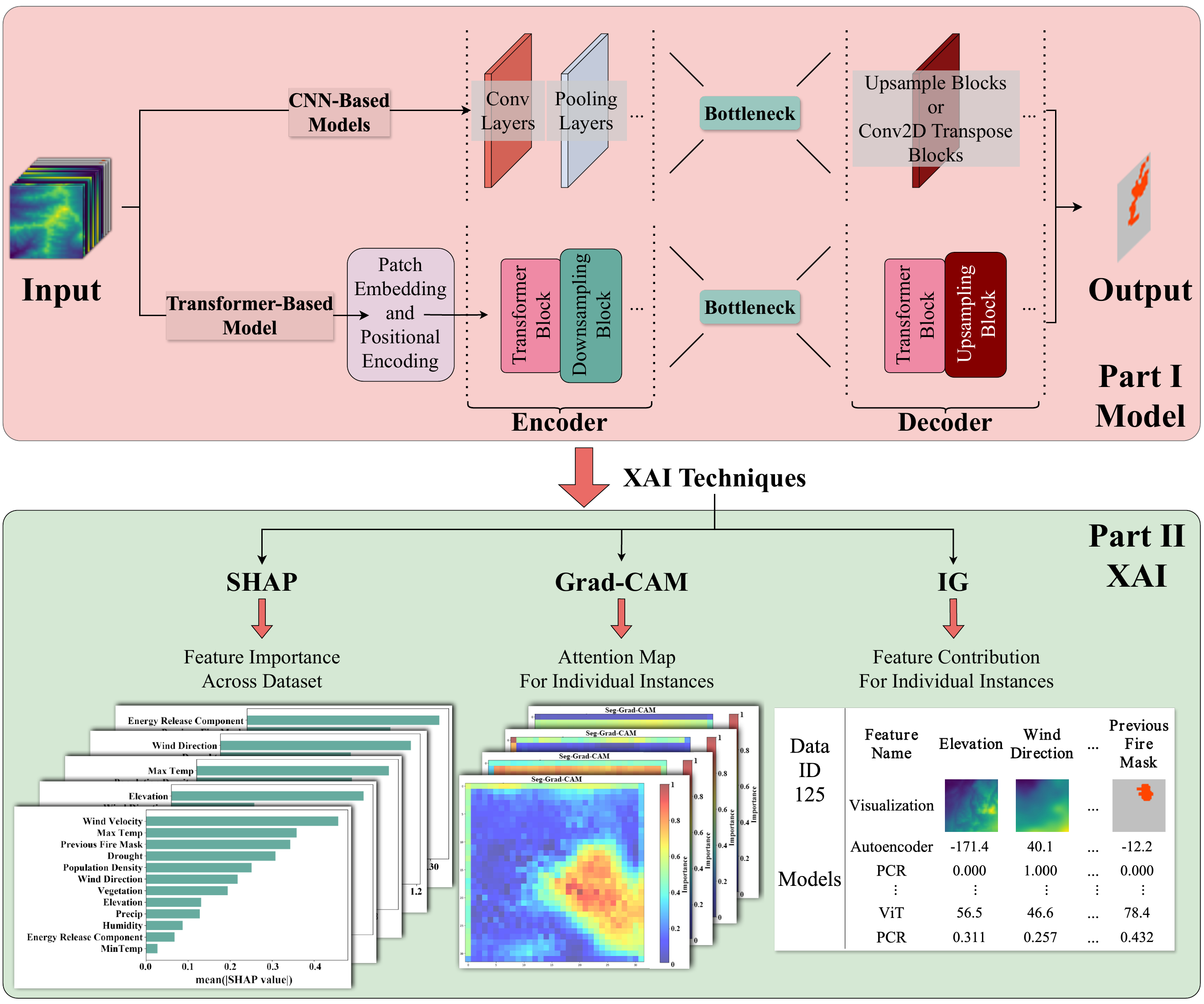}
    \caption{Workflow of this analysis. Starting with the input data \modified{(including meteorological data, multispectral data, terrain data, and so on)}, it is processed by two types of models: CNN-based models and a Transformer-based model. For all models, we further applied three explainability analysis techniques: SHAP, Grad-CAM, and IG, to explore and interpret the decision-making processes of the models. \modified{In the table of IG, 'Data ID 125' refers to the 125th sample in the dataset. PCR stands for Positive Contribution Rate, a metric indicating the contribution of a specific feature.}}
    \label{fig:flowchart}
\end{figure*}

Through SHAP, we quantify the impact of each environmental factors on model predictions across the entire dataset, laying a foundation for transparent model evaluation. Grad-CAM complements this by providing visual explanations that highlight critical areas within input images, thus allowing visual validation of the model's focus. IG extends this analysis by attributing the contributions of specific features to the prediction outcomes for individual samples in the dataset. Collectively, these methods enable us to pinpoint critical environmental factors, clarify their relative importance and interactions within the models. This approach not only augments the transparency and reliability of the four models we implemented but also fosters targeted enhancements in wildfire prediction capabilities. The overall workflow of this study is illustrated in Figure \ref{fig:flowchart}.

The primary objective of this research is to compare the efficiency of various deep learning models, specifically CNNs and Transformer-based model, in predicting wildfires using remote sensing data. These two categories of models possess inherently different characteristics, i.e., CNNs, such as autoencoders, ResNet, and UNet are good at capturing spatial hierarchies in imagery, while Transformers are efficient at modeling long-range dependencies and integrating context more comprehensively. We hypothesize that, these architectural differences will lead to varying performance in wildfire prediction tasks. Furthermore, by integrating explainability methods, such as SHAP values and Grad-CAM, we aim to offer valuable insights into the models' decision-making processes. Specifically, how they weigh and interpret different features in remote sensing data to arrive at their predictions. We expect this exploration to deepen our understanding of the models' predictive capabilities and the rationale behind their forecasts, thereby enhancing both the transparency and reliability of wildfire forecasting models.

\subsection{Related Works}
The evolution of wildfire prediction methods has significantly transitioned from reliance on traditional statistical methods to the incorporation of complex machine learning algorithms \cite{10069219, doi:10.1139/er-2020-0019}. Initially, predictions were primarily based on empirical models utilizing meteorological data and historical fire records, which despite their utility, often faced limitations such as time inefficiency and a lack of flexibility for parameter tuning \cite{perry1998current}. \changed{In addition to these empirical models, physical-based models have also been widely employed in wildfire prediction, such as The Wildland Fire Dynamics Simulator (WFDS) \cite{mell2007physics}, UoC-R \cite{zhou2005modeling}, and UoS \cite{asensio2002wildland}. These models, which are grounded in the fundamental chemistry and physics of combustion and fire spread, aim to simulate the behavior and dynamics of wildland fires more realistically \cite{sullivan2009wildland}.} The emergence of remote sensing technologies, such as satellite imagery and aerial photography, marked a considerable advancement by enriching the dataset available for analysis and improving the timeliness and reliability of wildfire predictions \cite{rashkovetsky2021wildfire, campbell2022application}.

Deep learning models, including CNNs like autoencoders, ResNet, UNet, and Transformer-based models, offer unique strengths in analyzing complex spatial and temporal patterns. \changed{Compared to traditional machine learning methods, deep learning approaches have been shown to achieve higher performance in wildfire prediction. For instance, in the comparison by \citeA{huot2021next}, autoencoders significantly outperformed traditional machine learning methods, with notable improvements in several metrics, demonstrating the ability of deep learning to capture more complex patterns and improve prediction accuracy.} Autoencoders are good at reducing dimensionality, denoising data, and discovering hidden features, making them effective for handling complex spatial and temporal data \cite{zhai2018autoencoder,cheng2022data}. ResNet's deep learning framework excels in learning detailed features through its extensive layers \cite{he2016deep}, while UNet combines context and localization, making it ideal for spatial tasks, such as the mapping of wildfire spread \cite{Hodges2019Wildland}. In contrast, Transformer-based model uses self-attention mechanisms, focusing on global data relationships rather than spatial proximity \cite{Pan2023Slide-Transformer:}, which is effective for understanding widespread environmental influences on wildfires. Additionally, Recurrent Neural Networks (RNNs) and Long Short-Term Memory networks (LSTMs) are commonly used for wildfire prediction due to their ability to handle sequential data and capture temporal dependencies. However, these models were not included in our comparative analysis because our focus is on evaluating the effectiveness of CNNs and Transformer models in handling the spatial complexities present in remote sensing data for wildfire prediction. 

Recent studies (\citeA{khanmohammadi2022prediction,zhong2023reduced,cheng2023generative}) have advanced the application of deep learning in wildfire prediction significantly, as summarized in \cite{xu2024wildfire}. For instance, deep learning models such as CNNs, RNNs and Transformers have been employed using satellite data to predict wildfire severity and danger, as reviewed by \citeA{DeepLearningSatellite2023}. Furthermore, \citeA{GlobalWildfire2024} proposed a model combining ConvLSTM networks with spatial feature extraction from U-Net and SKNet \cite{Li2019Selective} networks, effectively enhancing global wildfire danger predictions. Other notable efforts include the work by \citeA{9671680}, who collected SaskFire dataset and utilized ResNet, achieving a precision of 0.25 and a recall of 0.80 on a highly imbalanced dataset. Additionally, \citeA{Kondylatos2022Wildfire} explored CNNs and Transformer-based architectures, underscoring their effectiveness in forecasting wildfire danger. These differences highlight the necessity for comparative analysis to identify the most effective models for wildfire prediction.

\subsection{Contribution}
To address the 'black box' problem more comprehensively and achieve a deeper understanding of the decision-making process, XAI techniques such as SHAP, Grad-CAM, and IG are introduced. In the field of geoscience, researchers have acknowledged the significance of interpretability and have applied it in various domains, such as climate prediction \cite{Bommer2023Finding, Mamalakis2022Carefully} and drought forecasting \cite{Dikshit2021Interpretable}. Researchers have also attempted to incorporate explainability into wildfire scenarios. For instance, \citeA{FireXnet2023} designed FireXnet, achieving a 98.42\% test accuracy, which outperforming models like VGG16 and DenseNet201, and also integrated SHAP for explainability. Similarly, \citeA{qayyum2024shapley} employed a transformer encoder-based approach for wildfire prediction, using SHAP analysis to elucidate the connection between predictor variables and model performance. Overall, these studies highlight the growing precision and explanatory capabilities of deep learning models in wildfire prediction. However, the effectiveness and applicability of these methods in enhancing model transparency, especially within the context of remote sensing for wildfire prediction, have not been fully explored. For example, they often restrict explainability to general dataset-level analysis without further detailed examination, such as analyzing the outputs of each layer of the models and on each sample in the dataset, which could potentially offer deeper insights into the model's behavior and improve our understanding of its decision-making processes.

To address the identified research gaps and challenges in our study, we have undertaken a comprehensive approach that is outlined in several key areas. Firstly, we conducted a thorough quantitative analysis to compare the efficiency and performance of CNNs and Transformer models in predicting wildfires. This evaluation includes not just performance metrics but also efficiency aspects such as the number of parameters and GFLOPs, revealing distinct strengths and weaknesses of each model type. This comprehensive comparison serves as a foundation for understanding the trade-offs between model complexity and predictive accuracy. Secondly, we integrated advanced interpretability methods to investigate the decision-making processes behind the models' predictions. Our dual approach of analyzing the entire dataset through SHAP and assessing individual samples via Grad-CAM and IG clarifies how models prioritize and balance different environmental factors. This detailed analysis not only enhances model interpretability and transparency but also improves the credibility of wildfire prevention measures based on model predictions. \modified{Lastly, we balanced model performance and interpretability to provide valuable guidance for selecting the most suitable model for wildfire prediction, taking into account varying needs such as accuracy, recall, real-time performance, and computational resource demands.} The insights obtained from our study also provide a clear direction for future enhancements in wildfire prediction models, by identifying critical factors that influence model accuracy and interpretability.

In summary:
\begin{enumerate}
    \item We conducted a comprehensive analysis comparing models, including CNNs(a baseline Autoencoder proposed by \citeA{huot2021next}, ResNet, and UNet) and a Transformer-based model, all implemented from scratch, on their effectiveness and efficiency in wildfire prediction. Moreover, models such as UNet and Transformer-based Swin-UNet demonstrated superior performance compared to the baseline Autoencoder.
    \item We enhanced model transparency and interpretability through an integrative XAI approach incorporating SHAP, Grad-CAM, and IG. This contributes to advancing the application of XAI in predicting wildfires.
    \item We provided guidance for selecting suitable wildfire prediction models and outlined key areas for future research and enhancements.
\end{enumerate}

\changed{In Section \ref{methodology}, we detail the dataset utilized in this study, describe the architectures of the deep learning models employed, and outline the theoretical framework of the XAI methods applied. Section \ref{results} presents the evaluation metrics and offers a performance analysis based on the corresponding experimental results. Section \ref{xai} provides an in-depth interpretability analysis, examining each model feature by feature. Finally, the paper concludes with a summary and future directions in Section \ref{conclusion}.}

\section{Methodology} \label{methodology}

\subsection{Dataset Description}
Considering that both CNNs and Transformer-based models excel at processing complex spatial and temporal data, it is critical that our dataset is rich, multifaceted, and high-precision to fully leverage these characteristics of the models. Using primarily the 'Next Day Wildfire Spread' dataset \cite{huot2021next}, aggregated via Google Earth Engine \cite{gorelick2017google}, this study harnesses a comprehensive, multivariate collection of historical wildfire data across the United States. This dataset is unparalleled, integrating a decade's worth of satellite observations, standardized to a 1 km resolution, including previous and current fire masks from key sensors such as the Visible Infrared Imaging Radiometer Suite (VIIRS) and the Shuttle Radar Topography Mission (SRTM) \cite{VIIRS-VNP13A1, Farr2007}.

\remove{It includes 11 essential environmental variables, which provide a detailed view of fire events (18,545 in total) through daily snapshots, enabling precise modeling of fire spread patterns. Distinct from existing fire datasets that primarily focus on burn areas without offering comprehensive environmental context or the necessary 2-D, day-by-day progression for accurate fire spread analysis (such as FRY (Laurent et al., 2018), Fire Atlas (Andela et al., 2019), MOD14A1 V6 (Giglio \& Justice, 2015), GlobFire (Artes et al., 2019), VNP13A1 (Didan \& Barreto, 2018), and GRIDMET (Tavakkoli Piraliou et al., 2022)), our chosen dataset fills this gap by incorporating a wide array of variables that are critical for advanced predictive modeling. The 12 input features are 'Elevation', 'Wind direction', 'Wind velocity', 'Minimum temperature', 'Maximum temperature', 'Humidity', 'Precipitation', 'Drought', 'Vegetation', 'Population '' Density', 'Energy release component', and 'Previous fire mask'. The 12 features were selected by the original paper as they are important for predicting wildfire spread and are widely available from remote-sensing technologies (Huot et al., 2022). The output is the next day fire mask.}\changed{The dataset includes 12 essential input features that capture environmental, meteorological, and anthropogenic factors influencing wildfire behavior. These features are derived from various sources with different native spatial resolutions and temporal characteristics, all standardized to a 1 km resolution for consistency. Specifically, \textit{Elevation} is derived from SRTM data \cite{Farr2007} at a native resolution of 30 meters, providing critical terrain information. \textit{Wind direction} and \textit{Wind velocity} are obtained from the GRIDMET dataset \cite{rs14030672} at 4 km resolution, representing daily atmospheric conditions that affect fire spread and intensity. \textit{Minimum temperature}, \textit{Maximum temperature}, \textit{Humidity}, and \textit{Precipitation} are also sourced from GRIDMET\\ \cite{rs14030672} at 4 km resolution, capturing daily weather conditions influencing fuel moisture and fire ignition likelihood. The \textit{Drought} variable is derived from the U.S. Drought Monitor data, sampled every five days at 4 km resolution, integrating indicators such as precipitation, soil moisture, and streamflow to classify drought severity, which is crucial for assessing long-term dryness conditions that elevate wildfire risk. \textit{Vegetation} information is obtained from the Normalized Difference Vegetation Index (NDVI) provided by VIIRS \cite{VIIRS-VNP13A1} at a native resolution of 0.5 km, sampled every eight days, indicating fuel availability and condition vital for predicting fire spread. \textit{Population Density} is sourced from the Gridded Population of the World (GPWv4) dataset, updated every five years at 1 km resolution, serving as a proxy for human activity that may influence ignition rates and fire management practices. The \textit{Energy Release Component} (ERC), from the National Fire Danger Rating System, is available daily at 1 km resolution, representing potential energy release per unit area in the flaming front of a fire, reflecting fuel dryness and potential fire intensity. Lastly, the \textit{Previous fire mask} represents fire locations at time \( t \), obtained from MOD14A1 fire mask composites \cite{MOD14A1} at 1 km resolution, providing historical fire occurrence information essential for understanding fire progression.}

\changed{Each of these features is standardized to a spatial resolution of 1 km to ensure consistency across different data sources and facilitate effective integration into our models. The dataset spans from 2012 to 2020, utilizing daily snapshots or the most recent data available before the fire event (\( t \)), thereby ensuring that our models are informed by the conditions most likely to influence wildfire behavior. Most variables represent conditions immediately prior to the fire event, not long-term means. For example, meteorological variables and ERC are daily values capturing immediate conditions, while drought and vegetation indices use the most recent data before \( t \) to approximate current conditions. Population density and elevation are treated as static over the study period.}

This selection offers an opportunity to investigate wildfire spread with a level of detail and temporal resolution that is unmatched by other publicly available resources. \changed{Unlike existing fire datasets that primarily focus on burn areas without offering comprehensive environmental context or the necessary 2-D, day-by-day progression for accurate fire spread analysis (such as FRY \cite{laurent2018fry}, Fire Atlas \cite{andela2019global}, MOD14A1 V6 \cite{MOD14A1}, GlobFire \cite{artes2019global}, VNP13A1 \cite{VIIRS-VNP13A1}, and GRIDMET \cite{rs14030672}), our chosen dataset fills this gap by incorporating a wide array of variables critical for advanced predictive modeling. The output is the next day's fire mask, representing fire occurrences at time \( t+1 \).}

\begin{figure*}
    \centering
    \includegraphics[width=\textwidth]{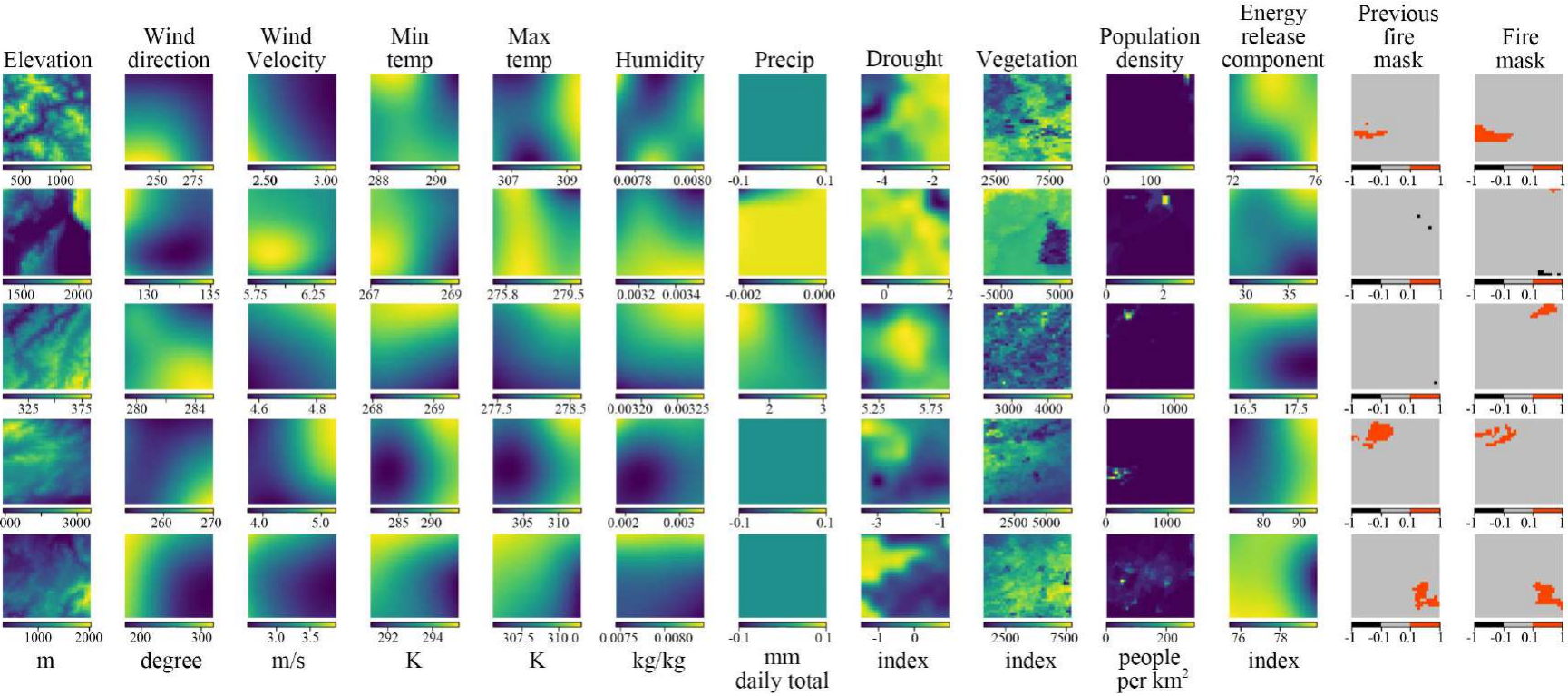}
    \caption{Visualized dataset \protect\cite{huot2021next}. \remove{The first 12 columns represent the input features, such as 'Elevation', 'Wind direction' and so on. The last column represents the label, where red indicates the presence of fire, gray signifies no fire, and black is used for uncertain labels, such as instances obscured by cloud coverage or other unprocessed data.} \changed{Each row represents a sample from the dataset, displaying all input features and the corresponding output. The first 12 columns represent the input features, such as 'Elevation', 'Wind direction', and so on. The last column represents the label, where red indicates the presence of fire, gray signifies no fire, and black is used for uncertain labels, such as instances obscured by cloud coverage or other unprocessed data.}}
    \label{fig:visualized_dataset}
\end{figure*}

The dataset is constructed from TFRecord files and configured with a batch size of 100, accommodating 12 input channels and 1 output channel. Some dataset visualizations are shown in Figure \ref{fig:visualized_dataset}. Our study involves a thorough data preprocessing stage, essential for the effective training of our models. Each feature is normalized based on predetermined statistics: minimum, maximum, mean, and standard deviation. To prepare the data, we used a random-cropping method, which ensures uniform-sized inputs for the model by extracting relevant sections from the images.

\subsection{Models}

\subsubsection{Autoencoder Architecture and Implementation}

\begin{figure*}[ht]
    \centering
    \includegraphics[width=\textwidth]{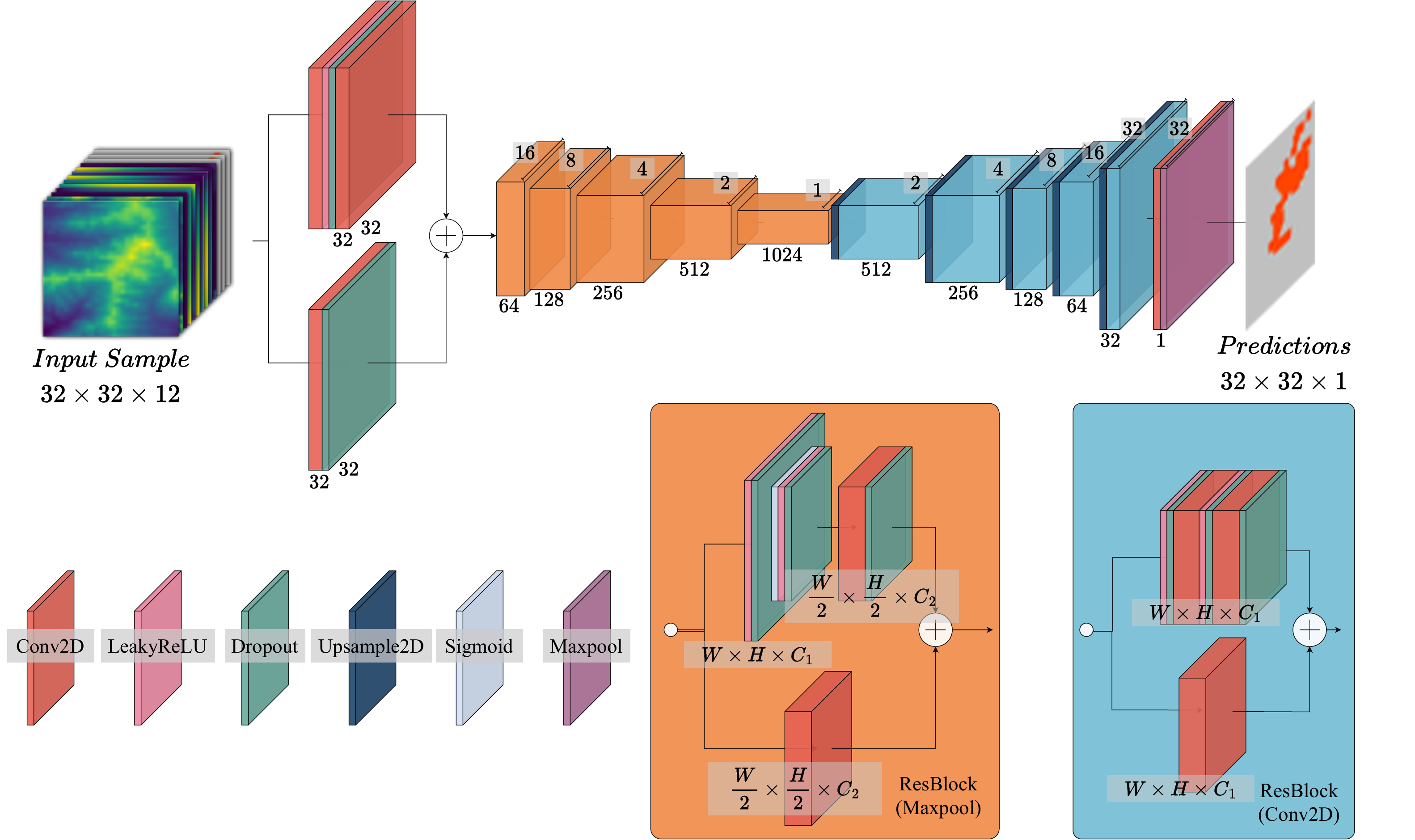}
    \caption{Detailed structure for baseline Autoencoder}
    \label{ae}
\end{figure*}

The autoencoder architecture, fundamental for dimensionality reduction in deep learning, was primarily inspired by \citeA{AutoencoderHinton}. Our model was adapted from the baseline model detailed in \cite{huot2021next}, which has 12 input channels and 1 output channel. The architecture of our model incorporates a sequence of convolutional layers, designed with dimensions following the sequence \([64, 128, 256, 256, 256]\). To complement the convolutional layers, each is succeeded by a pooling layer, specifically of size \(2 \times 2\), ensuring a structured reduction in spatial dimensions while retaining critical feature information. This configuration is strategically optimized to enhance the model's ability to capture and process the hierarchical spatial features essential for accurate wildfire prediction. The detailed structure of the Autoencoder can be seen in Figure \ref{ae}. 

\subsubsection{ResNet Architecture and Implementation}

\begin{figure*}[ht]
    \centering
    \includegraphics[width=\textwidth]{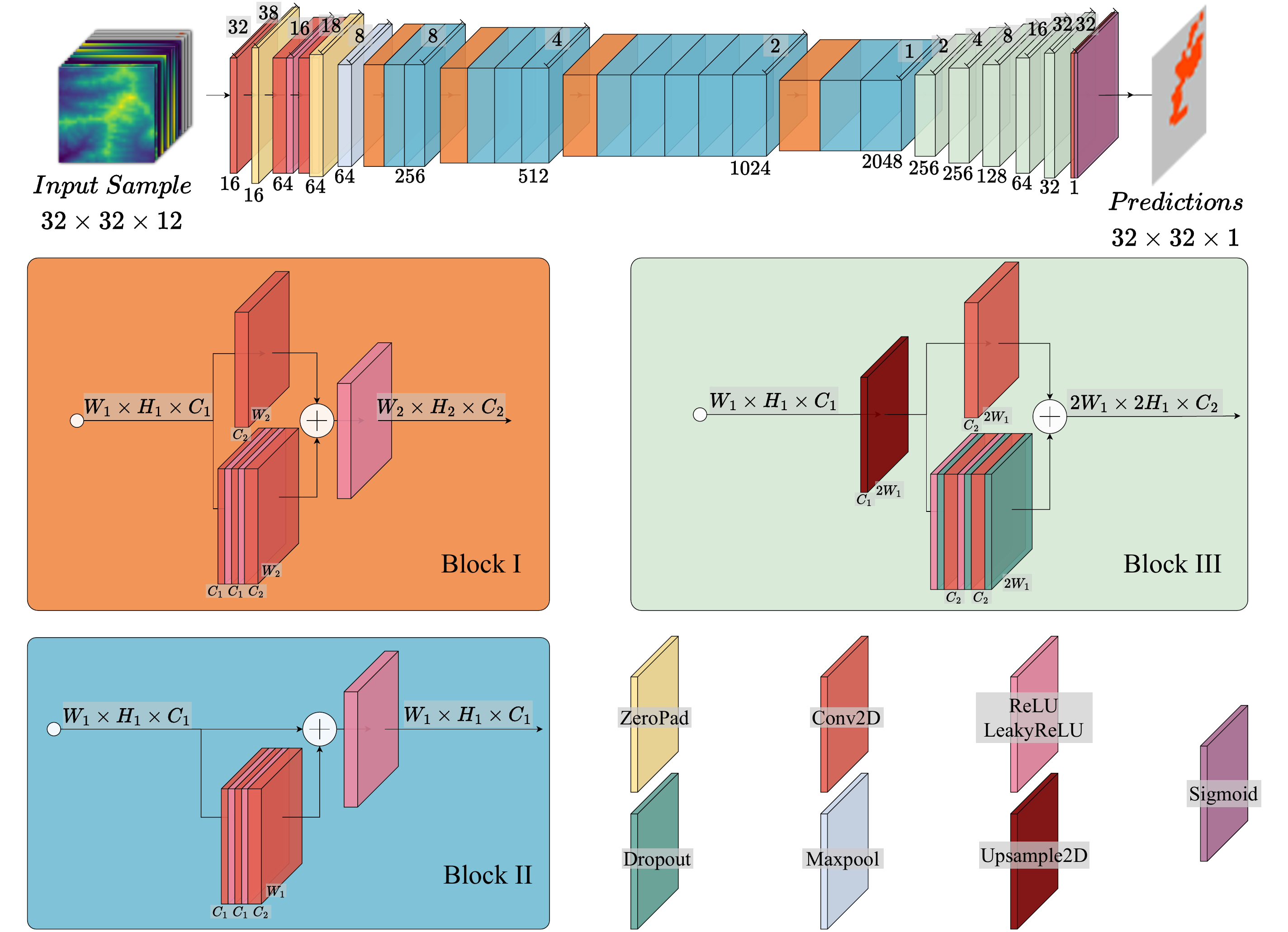}
    \caption{Detailed structure of ResNet, where skip connections are confined within individual blocks, which is different from the UNet that utilizes skip connections spanning from encoder to decoder.}
    \label{resnet}
\end{figure*}

The ResNet (Residual Network) architecture, a significant advancement in deep learning, is renowned for its deep structure that can run hundreds of layers. It addresses the vanishing gradient problem through 'skip connections', also known as 'residual connections', which allow the gradient to flow through the network without attenuation \cite{he2016deep}. ResNet's success in image recognition tasks has established it as a benchmark model in the field. In our study, we adapted the ResNet architecture, specifically using the ResNet50 variant, to predict wildfires from remote sensing data. To enhance our model's ability to extract features crucial for predicting wildfires, we added an additional convolutional layer before the ResNet encoder. This adjustment aims to improve the model's initial processing of input data. Further modifications include adopting the decoder structure from the existing baseline autoencoder model, integrating it with the ResNet encoder to form an efficient encoder-decoder setup. This setup is specifically designed to handle spatial features effectively, which are vital for accurate wildfire prediction. The decoder is structured to invert the process of the encoder, using reversed layers and pooling operations to reconstruct the output from encoded data. A customized architecture of the ResNet can be seen in Figure \ref{resnet}. The integration of ResNet with a decoder from an autoencoder model in our implementation is a novel approach. This combination leverages ResNet's powerful feature extraction capabilities and the decoder's efficient spatial reconstruction, making it highly suitable for the complex task of wildfire prediction from varied and intricate remote sensing data.

\subsubsection{UNet Architecture and Implementation}

\begin{figure*}[ht]
    \centering
    \includegraphics[width=\textwidth]{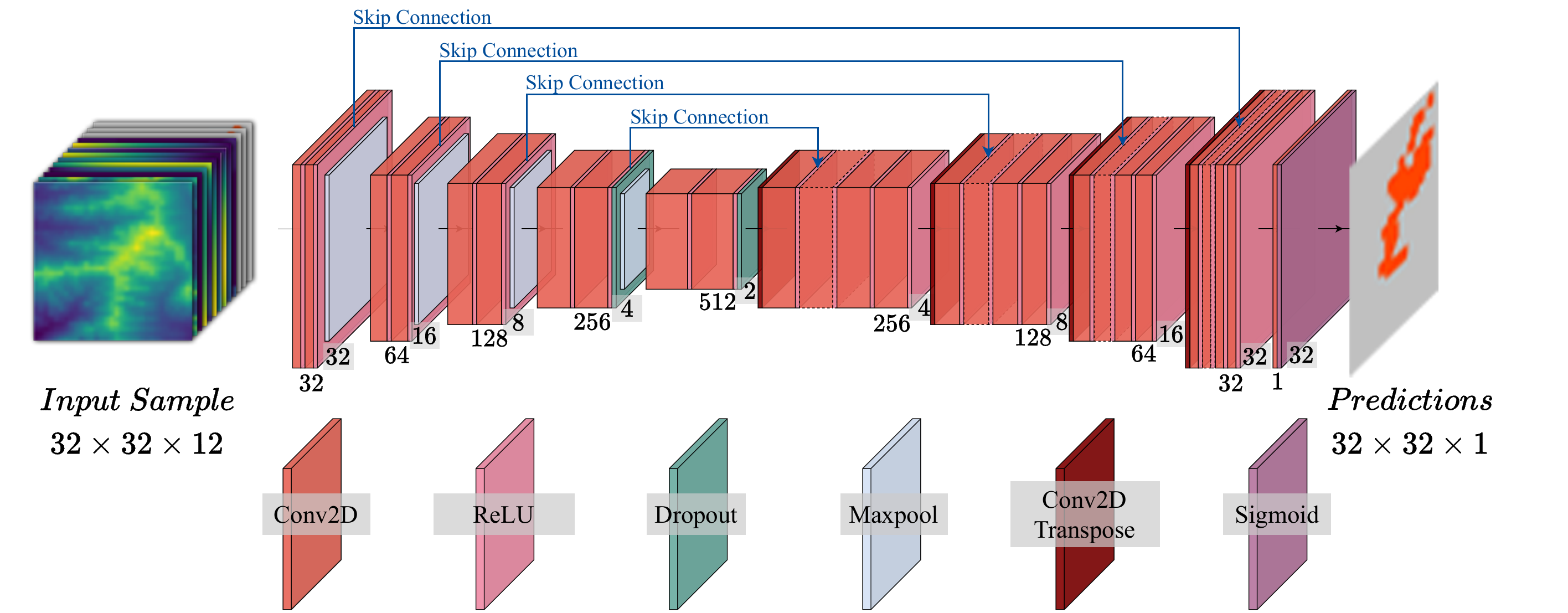}
    \caption{The detailed structure of UNet, featuring skip connections that extend from the encoder to the decoder, represented by blue lines in the illustration.}
    \label{unet}
\end{figure*}

The UNet architecture is distinguished by its U-shaped structure, featuring a contracting path for context capture and an expansive path for precise localization \cite{ronneberger2015u}. In particular, the UNet is distinguished in its ability for detailed segmentation tasks using a limited dataset. This allows us to accurately segment complex spatial patterns in satellite imagery and facilitates the identification of wildfire-prone areas. In our wildfire prediction study, we adapted the UNet model to process remote sensing data. Our model is initiated with an input layer designed to accommodate data dimensions of \(32 \times 32\) across 12 channels, setting the foundation for complex feature processing. The architecture progresses through a series of Convolutional 2D (Conv2D) layers, where the filter sizes gradually increase from 32 to 512. This escalation allows for a hierarchical extraction of features, ensuring a detailed understanding of the input data. Following each Conv2D layer, batch normalization is applied to stabilize the training process, addressing internal covariate shift and speeding up convergence. Spatial dimension reduction is achieved through max pooling in the contracting path, while the expanding path uses transposed convolutions for detailed spatial feature mapping. To take into account the risk of overfitting, dropout layers are placed in the deeper sections of the model. In our model, feature maps from the contracting path are precisely merged with those in the expansive path. The merging process preserves vital spatial details throughout the network, ensuring both the context and localization accuracy are enhanced for reliable wildfire prediction. The detailed structure of UNet can be seen in Figure \ref{unet}.

\subsubsection{Transformer-based Approaches Architecture and Implementation}

Transformer-based approaches introduce a novel method for image analysis by dissecting images into patches and processing them similarly to words in a sentence \cite{dosovitskiy2020image}. This technique allows these models to understand and connect different parts of an image globally, unlike traditional models that focus on local areas. In the context of wildfire prediction, it enables the model to detect subtle yet significant patterns across vast landscapes, such as changes in vegetation dryness or unusual temperature variations, which are key indicators of potential wildfire outbreaks. By capturing these comprehensive spatial relationships, transformer-based models can predict wildfires with higher accuracy, contributing to more effective monitoring and management of wildfire risks. Swin-UNet, an innovative adaptation of this architecture, is customized for segmentation tasks, enhancing the traditional UNet structure with Swin Transformer blocks \cite{SwinUNetCao}. This architecture is ideal for handling hierarchical features, a critical aspect in processing remote sensing data for wildfire prediction.

\begin{figure*}[ht!]
    \centering
    \includegraphics[width=\textwidth]{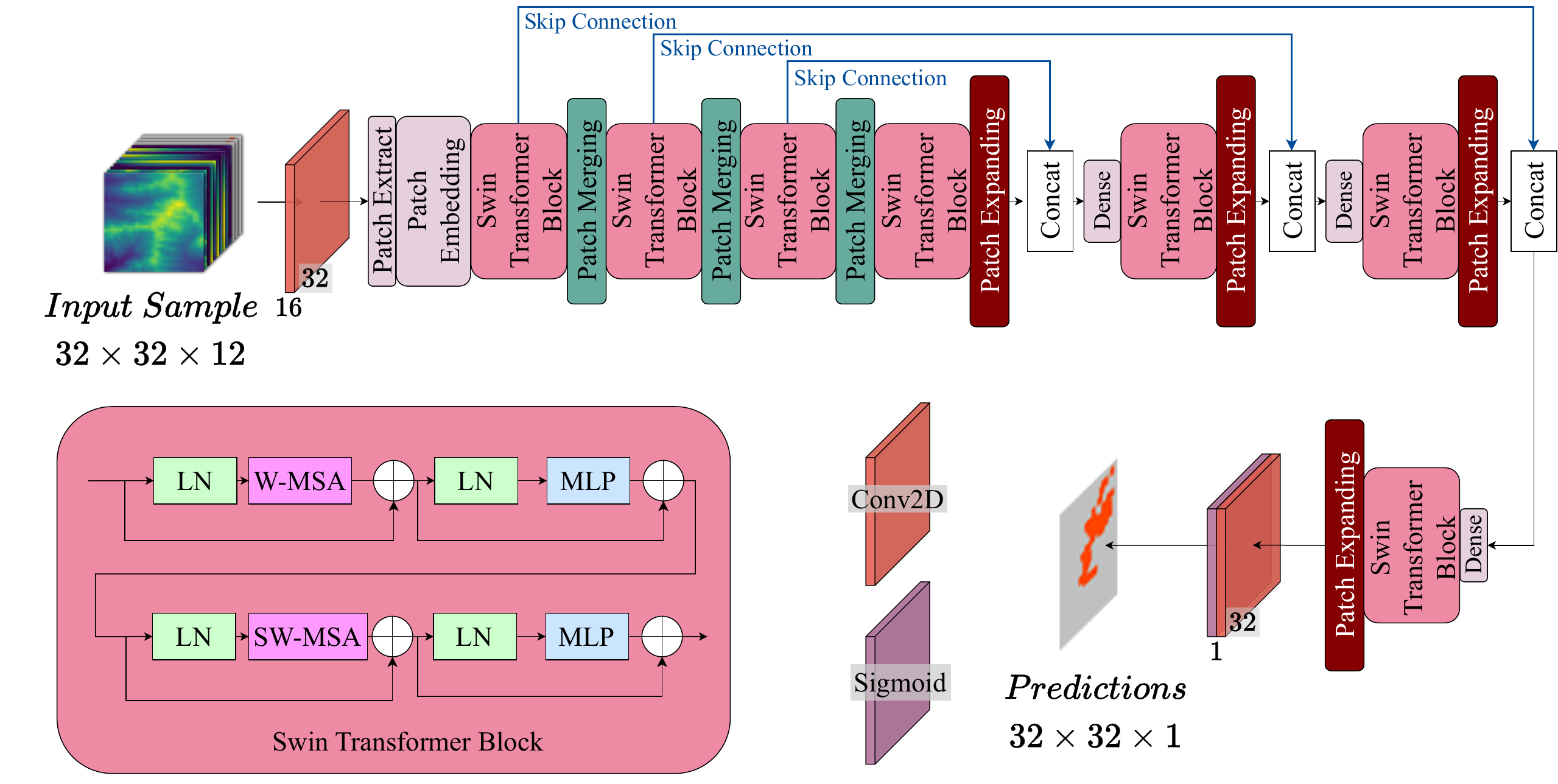}
    \caption{Detailed structure of Transformer-based Swin-UNet, featuring skip connections that extend from the encoder to the decoder, represented by blue lines in the illustration.}
    \label{vit}
\end{figure*}

In our Swin-UNet setup for wildfire prediction, we use a Swin Transformer for both the encoder and decoder parts. \modified{The model begins with a 16-channel convolutional layer to align with the initial convolutional layers of three other models, ensuring consistent heatmaps generated by Grad-CAM.} Following this, a downsampling block with 128 channels is employed to extract robust features. It has a depth of 4, including three down/upsampling levels and an extra bottom level for processing features at multiple scales. Each stage has two Swin Transformer blocks that use Window-based Multi-head Self-Attention (W-MSA) and Shifted Window-based Multi-head Self-Attention (SW-MSA) to capture both local and global details. W-MSA focuses on capturing spatial relationships within local windows, while SW-MSA shifts window positions to capture cross-window connections, enhancing global context integration. The $2 \times 2$ patch sizes help extract detailed image patches, which is important for capturing key details often missed by regular CNNs. The different attention heads and window sizes in the Swin Transformer blocks allow the model to analyze various spatial scales in an image, from large landscape features to small changes in vegetation or terrain. Layer Normalization (LN) is used to normalize the inputs across features for each layer, improving training stability and speed. The Multi-Layer Perceptron (MLP) introduces non-linearity through fully connected layers, helping the network learn complex features. \modified{During the upsampling process, the dense layers placed in conjunction with patch expanding layers increase the feature dimensions back to their original size.} This makes Swin-UNet very good for remote sensing data, combining effective spatial resolution and accurate segmentation tasks. The detailed architecture is shown in Figure \ref{vit}.

\subsubsection{Training Setup}
In the configuration of our model, both the input features and the target fire mask were established with spatial dimensions of \(32 \times 32\). The optimization process was executed using the Adam optimizer, which was configured with a learning rate of $\alpha = 0.0001$ and first and second-moment exponential decay rates set to \(\beta_1 = 0.9\), \(\beta_2 = 0.999\) \cite{kingma2015adam}.




To address the challenges posed by class imbalances and to focus more precisely on the classes most relevant for wildfire prediction, we adopted a custom \textit{masked weighted cross-entropy} loss function, as outlined in \cite{huot2021next}. Furthermore, to mitigate the risk of overfitting, we diverged from the original approach by incorporating an early stopping mechanism that concludes training if no improvement is observed over 30 epochs, in contrast to the original paper's use of a fixed 1000 epochs. While both studies employ the Adam optimizer, the precise optimization hyperparameters remain distinct and are customized in our study to address the challenges of wildfire prediction. The input features retain a resolution of $32 \times 32$ as established in the original framework, ensuring consistency in data representation. The Google Cloud computing T4 GPU is used for training, and the batch size is set to 100 for all the approaches.

The implementation of the deep learning models and interpretability techniques in this study used several key libraries. TensorFlow (2.13.0) and its Keras API were used for building and training the neural network models, including the Autoencoder, ResNet, UNet, and Vision Transformer (ViT). These tools provided robust capabilities for creating complex architectures and training them efficiently. NumPy (1.25.0) handled numerical computations and array operations, essential for data preprocessing and manipulation tasks. 

\subsection{Evaluation Metrics}
Given the nature of our task, which involves predicting wildfires from remote sensing data, we have chosen metrics that can provide a detailed understanding of model performance, particularly in the context of class imbalances and the critical importance of certain predictions. Our chosen metrics include:

\begin{enumerate}
    \item GFLOPs (Giga Floating Point Operations Per Second): This metric measures the computational complexity of the model in terms of the number of floating-point operations performed per second. GFLOPs provide a quantitative assessment of how demanding a model is in terms of computational resources, which is crucial for understanding its feasibility for deployment in real-time systems or on devices with limited processing power. Models with high GFLOPs might be more accurate but could suffer from longer inference times and higher power consumption, which are critical factors in applications like wildfire prediction where timely response is essential.
    \item Number of Parameters: This metric reflects the total count of trainable parameters within a model. A higher number of parameters generally indicates a more complex model that can capture more intricate patterns in the data. However, it also implies greater memory requirements and potential overfitting, especially in cases where data is scarce or noisy. Balancing the number of parameters is key to building efficient models that generalize well to new, unseen data without consuming excessive computational resources.
    \item AUC-PR (Area Under the Curve - Precision-Recall): This metric is particularly beneficial for our study due to its sensitivity to class imbalances. Unlike the more commonly used Area Under the Curve - Receiver Operating Characteristics (AUC), AUC-PR focuses on the relationship between precision (the proportion of true positive results among all positive predictions) and recall (the proportion of true positive results detected among all relevant samples). This focus makes AUC-PR more informative for datasets with a significant imbalance between classes, as is often the case in wildfire prediction, where the presence of fire is a relatively rare event compared to its absence. 
    \item Precision and Recall with Masked Class: To further customize our evaluation to the specific challenges of our dataset, we have implemented custom versions of precision and recall metrics that specifically exclude uncertain labels, which are represented as '-1' in our dataset. This approach ensures that our evaluation metrics only consider relevant classes for wildfire prediction, enhancing the focus on accurately predicting fire presence or absence without being disturbed by masked regions in the data.
    \item AUC: While AUC is less sensitive to class imbalance than AUC-PR, it remains a valuable metric for evaluating overall model performance. AUC measures the ability of the model to distinguish between classes across all thresholds, providing a comprehensive overview of model effectiveness.
    \item Confusion Matrix: This metric visualizes the performance of a classification model by displaying the true positives, true negatives, false positives, and false negatives in a matrix format. It helps in understanding the types of errors made by the model and the classes that are most often misclassified. Including a confusion matrix in our evaluation allows for a deeper analysis of how well the model performs specifically in distinguishing between the presence and absence of wildfires, considering the actual and predicted classifications.
    \item Pearson Correlation Coefficient: This statistical measure assesses the linear relationship between two features, providing insights into the degree of correlation between the predicted and actual values. A higher Pearson correlation coefficient indicates a stronger direct linear relationship. \changed{In this study, the Pearson correlation is calculated based on the the raw data values of the corresponding features.}
    \item Structural Similarity Index Measure (SSIM): This metric evaluates the visual impact of changes between two images, which is particularly useful in tasks that involve image processing or comparison. SSIM provides a more perceptual-based measure that compares the structural information in the images, offering a different perspective than purely pixel-based differences. \changed{In this study, the SSIM is calculated based on the the raw data values of the corresponding features.}
\end{enumerate}

By leveraging these metrics, we aim to gain a thorough understanding of our models' predictive capabilities, with a particular emphasis on their ability to recognize and accurately predict wildfire occurrences.

\subsection{Interpretability Techniques}

In this study, we incorporated SHAP, Grad-CAM, and IG to achieve a better understanding of the implemented models. Before investigating the results of those techniques, it's essential to understand the underlying logic and mathematical principles of them.

\subsubsection{SHAP}
\remove{SHAP, rooted in game theory and introduced by Shapley in 1953 (Shapley, 1953), has become critical in enhancing the interpretability of complex machine learning models (Lellep et al., 2022; Z. Li, 2022). It quantifies feature importance through the average marginal contribution of each feature, ensuring equitable attribution to the model's output by considering all feature combinations. Lundberg and Lee (2017) extended SHAP to enabled its application to more models, including tree-based models like XGBoost and deep learning architectures.} \newchanged{Shapley values}\changed{, rooted in game theory and originally introduced by Shapley in 1953 \cite{shapley1953value}, provides the mathematical foundation for quantifying the importance of features through the average marginal contribution of each feature to the model's output. This ensures equitable attribution by considering all feature combinations. Later, \citeA{lundberg2017unified} extended these concepts to machine learning explainability, specifically adapting SHAP for complex models, including tree-based algorithms like XGBoost and deep learning architectures, thus establishing SHAP explanations as a modern interpretability tool \cite{lellep_prexl_eckhardt_linkmann_2022, LI2022101845}.}

The Shapley value which quantifies average contribution for a feature \( i \) in a model is mathematically defined as:
\begin{linenomath*}
\begin{equation}
    \scalebox{1}{\( \displaystyle
    \phi_i (v) = \sum_{S \subseteq N \setminus \{i\}} \frac{|S|! \cdot (|N| - |S| - 1)!}{|N|!} \left ( v (S \cup \{i\}) - v (S) \right)
    \)}
\end{equation}
\end{linenomath*}
where \( \phi_i(v) \) is the Shapley value for feature \( i \). Here, \( S \) represents a subset of features excluding feature \( i \). \( N \) is the set of all features in the model, with \( |N| \) representing the total number of features. Additionally, \( |S| \) indicates the number of features in subset \( S \). The expression \( v(S \cup \{i\}) - v(S) \) calculates the marginal contribution of feature \( i \) when added to the subset \( S \). Function \( v(S) \) is the prediction of the model using the features in subset \( S \), whereas \( v(S \cup \{i\}) \) is the prediction using features in \( S \) along with feature \( i \). The coefficient \(\frac{|S|! \cdot (|N| - |S| - 1)!}{|N|!}\) serves as a weighting factor that accounts for the number of permutations of feature subsets, ensuring a fair distribution of contribution among all features. The Shapley value therefore refers to the average contribution of each feature to the predictive outcome of a model, and also allowing us to understand the importance of a feature. \remove{The use of the Shapley value is important in applications such as wildfire prediction because it can account for correlations among features, effectively handling the common issue of feature correlation in environmental data. This enables us to accurately identify and understand which features play key roles in model predictions, even in the presence of complex interactions among features.}  \changed{In this study, we utilized Gradient Explainer from the SHAP library to compute Shapley values efficiently, leveraging its compatibility with diverse deep learning architectures, including convolutional neural networks and transformers, as well as its adaptability to high-dimensional inputs such as our dataset.} To establish a baseline model and further analyze the SHAP differences between tree-based models and deep learning models, we also employed XGBoost, imported from the ‘xgboost’ library, on the same dataset with Tree Explainer.

\subsubsection{Grad-CAM}

Grad-CAM, crucial for enhancing the interpretability of deep learning models, especially CNNs, was introduced by \citeA{selvaraju2017grad}. It provides a heatmap of influential regions within images for specific class predictions. This paper seeks to leverage Grad-CAM to visualize and compare the attention mechanisms inherent in the different approaches we have explored, namely Autoencoder, ResNet, UNet, and Transformer-based Swin-UNet. Consider the selected feature maps \(\{A^k\}_{k=1}^{K}\) (\(K\) kernels from the final convolutional layer of a classifier), and let \(y^c\) be the logit corresponding to a specific class \(c\). Grad-CAM computes the mean of the gradients of \(y^c\) across all \(N\) pixels ((identified by coordinates \(u, v\)) within each feature map \(A^k\), resulting in a weight \(w_k^c\) that denote its importance. The heatmap
\begin{linenomath*}
\begin{equation}
    L_{\text{Grad-CAM}}^c = \text{ReLU} \left ( \sum_k w_k^c A^k \right) \quad \text{with} \quad w_k^c = \frac{1}{N} \sum_{u,v} \frac{\partial y^c}{\partial A^k_{u, v}}
\end{equation}
\end{linenomath*}
is then produced by summing the feature maps with the respective weights, followed by the application of a pixel-wise ReLU operation to set negative values to zero. This process ensures that only regions positively influencing the class \(c\) decision are emphasized. While a classification network outputs a single class probability distribution for each input image \(\mathbf{x}\), a segmentation model (such as our wildfire segmentation approaches) assigns logits \(y_{i, j}^c\) to every pixel \(x_{i, j}\) for each class \(c\). Thus, \citeA{vinogradova2020towards} modified the original equation by substituting \(y^c\) with \(\sum_{ (i,j) \in M} y_{i, j}^c\) , where \(M\) denotes the set of pixel indices of interest within the output mask:
\begin{linenomath*}
\begin{equation}
    L_{\text{SEG-Grad-CAM}}^c = \text{ReLU} \left ( \sum_k w_k^c A^k \right) \quad \text{with} \quad w_k^c = \frac{1}{N} \sum_{u,v} \frac{\partial \sum_{ (i,j) \in M} y_{i, j}^c}{\partial A^k_{u, v}}
\end{equation}
\end{linenomath*}
\remove{It allows SEG-Grad-CAM capable for a fire segmentation task, since it can deal with logits of all fire areas, not just a single logit (like the output of a classification model).}\changed{SEG-Grad-CAM is well-suited for fire segmentation tasks because it processes logits from all fire-affected regions in an image, allowing for detailed spatial localization of fire areas. Unlike classification models that produce a single logit indicating the presence of fire, SEG-Grad-CAM can generate comprehensive activation maps that highlight each specific area where fire is detected.} This method significantly advances our comprehension of how models interpret remote sensing data at the pixel level, a key for environmental monitoring. Additionally, we employ SEG-Grad-CAM to approximate the visualization of our dataset's original 12 features, directing our attention to the first Conv2D layer. This strategy captures low-level features, ensuring a closer resemblance to the original data's visual characteristics through a channel-specific analysis.

\remove{This approach helps each model generates separate heatmaps for their first convolutional layer's 16 channels and a combined heatmap (also for the first convolutional layer) for a comprehensive view of all channels' impact on model predictions.}\changed{This approach generates individual heatmaps for each of the 16 channels in the first convolutional layer of each model, as well as a combined heatmap that aggregates the contributions of all channels. This dual visualization provides a comprehensive understanding of how each channel and the collective set of channels influence the model's predictions.} Our method, integrating individual and combined attention heatmaps, facilitates a thorough understanding of the model's information processing and decision-making, thereby identifying crucial feature regions for accurate predictions.

\subsubsection{Integrated Gradients}
\remove{Grad-CAM outperforms at identifying features with distinct visual characteristics, such as 'Elevation', 'Drought', 'Vegetation', 'Population density', and 'Previous Fire Mask' (PFM), but fails to express abstract features such as 'Min temp' and 'Humidity' effectively. IG, introduced by Sundararajan et al. (2017), complements Grad-CAM by providing a quantitative analysis of abstract features which lack clear visual patterns.}
\changed{Grad-CAM effectively highlights features with distinct visual patterns, such as 'Elevation', 'Drought', 'Vegetation', 'Population density', and 'Previous Fire Mask' (PFM), by mapping their contributions onto heatmaps. However, for abstract or less visually discernible features, such as 'Min temp' and 'Humidity', Grad-CAM struggles to provide meaningful insights due to the lack of clear spatial characteristics. In these cases, IG complements Grad-CAM by providing quantitative measures of feature importance\cite{sundararajan2017axiomatic}, enabling a more comprehensive understanding of both visually apparent and abstract features.}
IG quantifies the contribution of input features by calculating the gradient of the model's output relative to each input feature along a linear path from a baseline to the actual input. The attribution for each input pixel value \( x_i \) against a baseline pixel value \( x'_i \) is calculated by:
\begin{linenomath*}
\begin{equation}
    \scalebox{1}{\( \displaystyle
    IG_i = (x_i - x'_i) \times \int_{\lambda=0}^{1} \frac{\partial F (x' + \lambda \times (x - x'))}{\partial x_i} d\lambda
    \)}
\end{equation}
\end{linenomath*}
where \( F \) represents the model and \( \lambda \) varies from 0 to 1. This integral effectively captures the accumulated gradient contributions along the path from the baseline to the actual input, proportionally scaled. \changed{For our experiments, we used a zero baseline ($x'_i=0$) for all features, which is a common choice in XAI methods to represent the absence of input signals.}

In wildfire segmentation from remote sensing data, IG enables precise analysis at the pixel level, identifying features crucial for fire prediction. Employing IG, we obtain a $12\times32\times32$ attribution array per sample, reflecting the contributions of 12 features. By aggregating the values within each feature's matrix, we derive 12 comprehensive scores, quantifying the overall importance of each feature in the model's predictions. However, the analysis remained challenging since scores hardly reflect the degree of feature importance. Therefore, we calculated the positive contribution ratio (PCR) for each feature, revealing the proportion of each feature's positive contribution compared to others. For the \(i\)-th feature:
\begin{linenomath*}
\begin{equation}
    \scalebox{1}{\( \displaystyle
    \text{PCR}_i = \frac{\max (\gamma_i, 0)}{\sum_{i=1}^{12} \max (\gamma_i, 0)}
    \)}
\end{equation}
\end{linenomath*}
where \(\gamma_i\) represents the feature contribution of the \(i\)-th feature, for \(i = 1, 2, ..., 12\). This method allows us to discern key features determining fire presence, enhancing our model's interpretability and guiding future model development. Integrating IG with Grad-CAM offers a thorough framework for analyzing feature contributions in wildfire prediction, enriching our understanding of the model's decision-making process.

\subsection{Experimental Design}
Our research methodology adopts a systematic experimental framework and examines the explainability and interpretability of four advanced models: Autoencoder, ResNet, UNet, and Transformer-based Swin-UNet. This framework is structured to ensure the reproducibility and reliability of models, and also to test the performance of the models under different training situations. In particular, we test the models under the following two variations:
\begin{enumerate}
    \item Variation in Random Seed
    
    To ensure the reproducibility and reliability of these models, an experiment will be conducted in which each model undergoes training four times. The training, validation, and test datasets are fixed, but different random seeds are selected for initialize each training session. This choice is designed to probe the models' performance consistency, effectively controlling for the variability introduced by stochastic initialization.

    \item Variation in Dataset Proportion

    To investigate how the volume of training set impacts our models' performance, the validation set and test set were fixed, and each model will systematically be trained across a spectrum of training set proportions: 10\%, 25\%, 50\%, 75\%, and 100\%. This tiered training approach is instrumental in evaluating the models' adaptability and performance across varying data availability. This experiment helps us distinguish models that are more adept at learning from limited data from those that require larger datasets to achieve optimal performance, offering insights into real-world wildfire prediction scenarios given varying data availability.
\end{enumerate}


\section{Results and Discussions} \label{results}

\begin{figure}[ht!]
    \centering
    \begin{minipage}{0.45\textwidth}
        \centering
        \includegraphics[width=\textwidth]{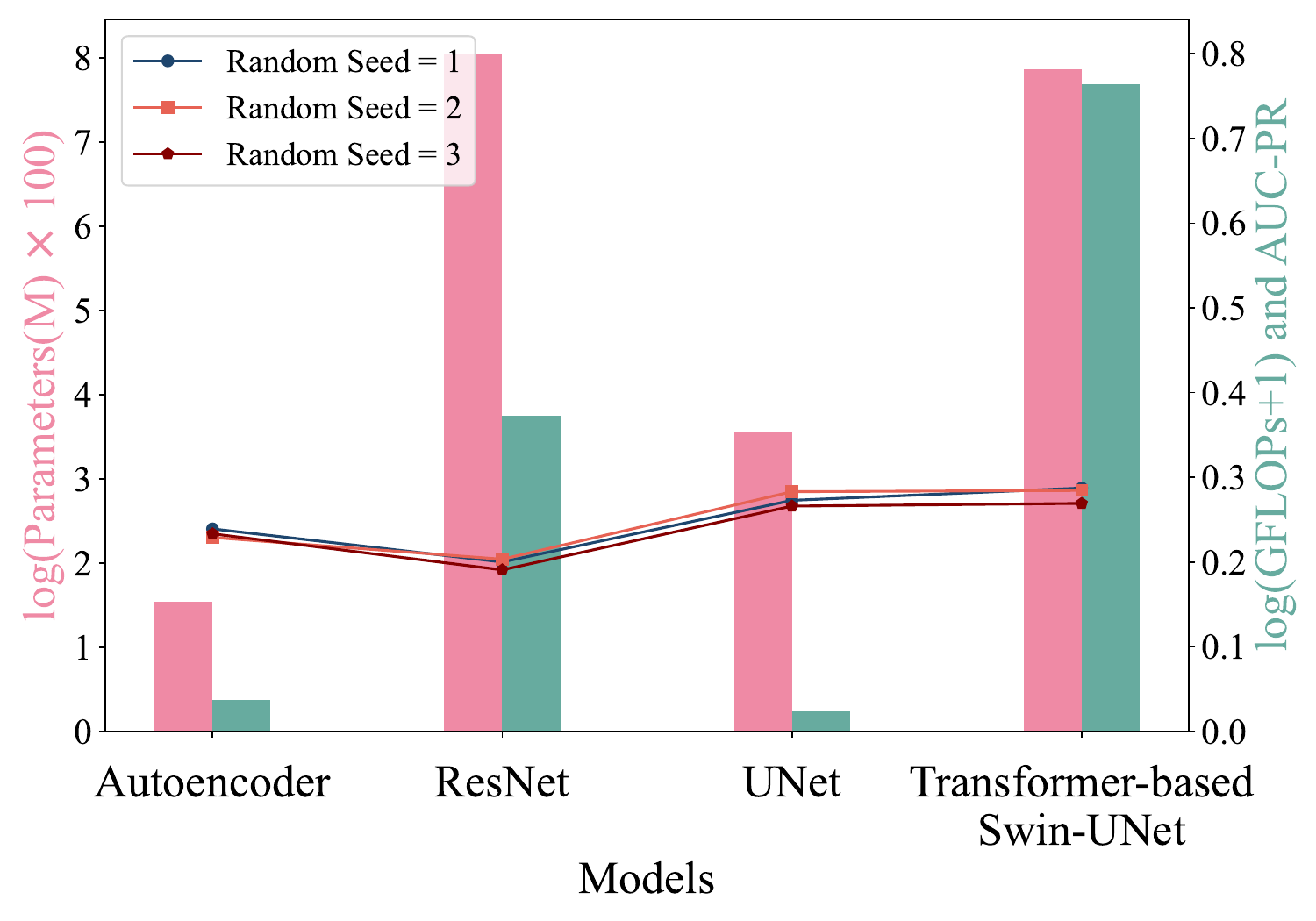}
        \caption{Comparison of models trained with different random seed. Each line represents the AUC-PR for each model with different initialization.}
        \label{fig: random seed comparison}
    \end{minipage}\hfill
    \begin{minipage}{0.45\textwidth}
        \centering
        \includegraphics[width=\textwidth]{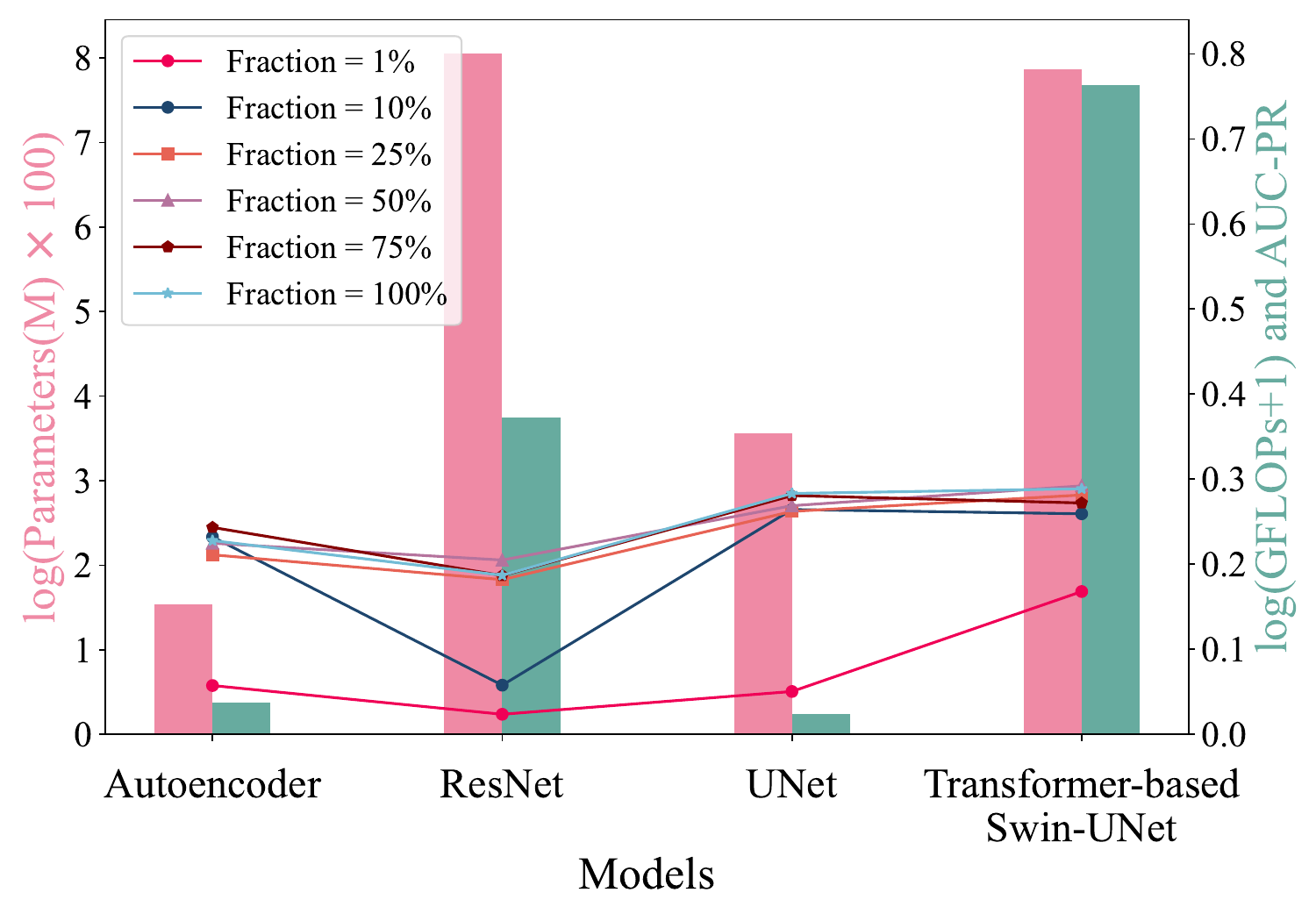}
        \caption{Comparison of models trained with different fraction of the dataset. Each line represents the AUC-PR for each model with different training volume.}
        \label{fig: fraction comparison}
    \end{minipage}
\end{figure}

This section analyzes performance metrics for various models under distinct conditions, using Tables \ref{Table: Random Seed} and \ref{Table: Fractions} to present metrics from training with different random seeds and dataset volumes. Figure \ref{fig: random seed comparison} visualizes the impact of initial seed variability on AUC-PR, parameters, and GFlops, while Figure \ref{fig: fraction comparison} illustrates how dataset volumes affect models performance. In Tables \ref{ig-grad-5} and \ref{ig-grad-15}, 'Predict Fire Mask' shows the prediction result according to each model. These visuals offer clear insights into the differences between models and the influences of seed settings and data size.

\begin{table}[ht!]
\centering
\begin{tabular}{lcccc}
\hline
                        & \multicolumn{1}{l}{Autoencoder} & \multicolumn{1}{l}{ResNet} & \multicolumn{1}{l}{UNet} & \multicolumn{1}{l}{Transformer-based Swin-UNet} \\ \hline
GFlops                  & 0.038                           & 0.452                      & 0.024                    & 1.146                   \\
Number of Parameters (M) & 0.046                           & 31.332                     & 0.353                    & 25.995                  \\ \hline
\end{tabular}
\caption{GFlops and Number of Parameters for 4 models}
\label{Table: parameters}
\end{table}

\begin{table}[ht!]
\centering
\resizebox{0.95\textwidth}{!}{
\begin{tabular}{cccccc}
\hline
Model & AUC $\pm$ error rate & AUC-PR $\pm$ error rate & Precision $\pm$ error rate & Recall $\pm$ error rate \\ \hline
Autoencoder & 0.8457 $\pm$ 0.64\% & 0.2338 $\pm$ 2.18\% & 0.3418 $\pm$ 3.83\% & 0.2551 $\pm$ 13.14\% \\
ResNet & 0.8274 $\pm$ 2.39\% & 0.1980 $\pm$ 3.28\% & 0.2985 $\pm$ 10.62\% & 0.2368 $\pm$ 35.98\% \\
UNet & 0.8463 $\pm$ 2.13\% & 0.2739 $\pm$ 3.14\% & 0.3464 $\pm$ 2.42\% & \textbf{0.3859} $\pm$ 2.72\% \\
\makecell{Transformer-based \\ Swin-UNet} & \textbf{0.8637} $\pm$ 0.44\% & \textbf{0.2803} $\pm$ 3.46\% & \textbf{0.3686} $\pm$ 1.19\% & 0.3470 $\pm$ 5.53\% \\ \hline
\end{tabular}
}
\caption{Performance metrics evaluated on the Test Set, averaged over 3 random seeds. \changed{The error rates reflect the variability of model performance across different initializations, with larger error rates indicating greater instability relative to the model's own metrics.}}
\label{Table: Random Seed}
\end{table}

\begin{table}[ht]
\centering
\begin{tabular}{ccllll}
\hline
Fraction & Model & \multicolumn{1}{c}{AUC} & \multicolumn{1}{c}{AUC-PR} & \multicolumn{1}{c}{Precision} & \multicolumn{1}{c}{Recall} \\ \hline
\multirow{4}{*}{10\%} & Autoencoder & 0.8477 & 0.2321 & 0.2788 & 0.3903 \\
 & ResNet & 0.7146 & 0.0578 & 0.2471 & 0.0004 \\
 & UNet & 0.8340 & \textbf{0.2643} & \textbf{0.3381} & 0.3875 \\
 & Transformer-based Swin-UNet & \textbf{0.8520} & 0.2593 & 0.3159 & \textbf{0.4125} \\
\multirow{4}{*}{25\%} & Autoencoder & 0.8527 & 0.2111 & 0.2858 & 0.3367 \\
 & ResNet & 0.8105 & 0.1819 & 0.2662 & 0.2755 \\
 & UNet & 0.8467 & 0.2621 & 0.3211 & 0.3972 \\
 & Transformer-based Swin-UNet & \textbf{0.8699} & \textbf{0.2814} & \textbf{0.3412} & \textbf{0.4069} \\
\multirow{4}{*}{50\%} & Autoencoder & 0.8343 & 0.2247 & 0.3120 & 0.3142 \\
 & ResNet & 0.8454 & 0.2049 & 0.2862 & 0.2835 \\
 & UNet & 0.8469 & 0.2688 & 0.3389 & 0.3745 \\
 & Transformer-based Swin-UNet & \textbf{0.8726} & \textbf{0.2921} & \textbf{0.3644} & \textbf{0.3917} \\
\multirow{4}{*}{75\%} & Autoencoder & 0.8608 & 0.2433 & 0.3359 & 0.2900 \\
 & ResNet & 0.8387 & 0.1867 & 0.2718 & 0.2658 \\
 & UNet & 0.8586 & \textbf{0.2807} & \textbf{0.3604} & \textbf{0.3752} \\
 & Transformer-based Swin-UNet & \textbf{0.8696} & 0.2720 & 0.3544 & 0.3493 \\
\multirow{4}{*}{100\%} & Autoencoder & 0.8475 & 0.2278 & 0.3172 & 0.2883 \\
 & ResNet & 0.8273 & 0.1870 & 0.3006 & 0.1644 \\
 & UNet & 0.8337 & 0.2832 & \textbf{0.3584} & \textbf{0.4040} \\
 & Transformer-based Swin-UNet & \textbf{0.8707} & \textbf{0.2887} & 0.3540 & 0.3880 \\ \hline
\end{tabular}
\caption{Training results evaluated on Test Set for different fractions of dataset.}
\label{Table: Fractions}
\end{table}

\subsection{Fundamental Performance Metrics}
AUC and AUC-PR are critical indicators of model performance, in Figure \ref{fig: random seed comparison} and Figure \ref{fig: fraction comparison}, the Transformer-based Swin-UNet and UNet models generally outperform Autoencoder and ResNet in both AUC and AUC-PR scores, indicating their superior ability to correctly predict wildfire events. The closeness in AUC scores among all models suggests they are generally comparable in distinguishing between fire and non-fire areas. However, the larger variance in AUC-PR scores highlights that the Transformer-based Swin-UNet and UNet are particularly better at balancing precision and recall, crucial for imbalance scenarios such as wildfire.
\begin{figure}[htbp]
    \centering
    \setlength{\abovecaptionskip}{5pt}
    \setlength{\belowcaptionskip}{5pt}
    \begin{minipage}{0.7\linewidth}
        \includegraphics[width=\linewidth]{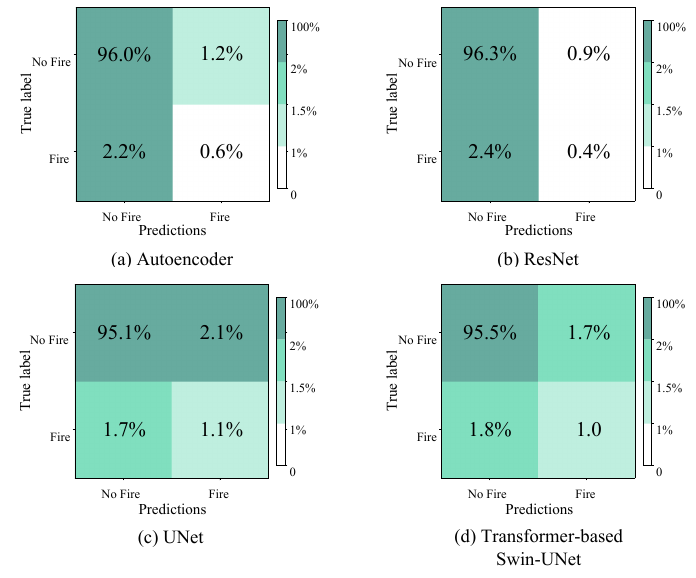}
        \caption{Confusion matrices for four models. The color scale is nonlinear due to the imbalance in the dataset.}
        \label{confusion_matrices}
    \end{minipage}
\end{figure}
To further analyze the practical implications of these performance metrics, we examine the confusion matrices of each model along with Precision and Recall. 

This section focuses specifically on the dataset imbalance and its impact on model performance. Due to a significantly higher number of negative (i.e., without fire) samples compared to positive (i.e., with fire) samples, this imbalance is crucial for evaluating model performance. As shown in Figure \ref{confusion_matrices}, all models exhibit a significant discrepancy with True Negatives (TN) far outnumbering True Positives (TP), indicating a dataset bias towards 'No Fire' instances. Transformer-based Swin-UNet and UNet models perform better in identifying TP instances, reducing the occurrence of False Negatives (FN), which explains why these two models typically have higher Recall in Tables \ref{Table: Random Seed} and \ref{Table: Fractions}. \modified{As shown in the 'Predict Fire Mask' part of Tables \ref{ig-grad-5}, \ref{ig-grad-15}, \ref{ig-grad-21}, \ref{ig-grad-29}, and \ref{ig-grad-125}, UNet and Transformer-based Swin-UNet usually have a wider range of predictions compared to the Autoencoder and ResNet. This broader range helps UNet and Transformer-based Swin-UNet to identify more TP instances, which in turn reduces the occurrence of FN. Specifically, the wider prediction range of UNet and Transformer-based Swin-UNet allows these models to capture more subtle variations in the data that might be missed by the Autoencoder and ResNet. In contrast, the Autoencoder and ResNet tend to have a more conservative prediction range, which often leads to higher FN rates. This conservatism means that FN instances in these models are more likely to appear near the boundary of the TP region, where the prediction confidence is lower. As a result, UNet and Transformer-based Swin-UNet typically have higher Recall, as shown in Tables \ref{Table: Random Seed} and \ref{Table: Fractions}.}

On the other hand, the UNet and Transformer-based Swin-UNet might be more sensitive to positive samples in handling imbalanced datasets, but they also produce more False Positives (FP), as illustrated in Figure \ref{confusion_matrices}, where UNet and Transformer-based Swin-UNet show more FP than the other two models. However, the other two models' performance in recognizing TP is significantly worse than that of UNet and Transformer-based Swin-UNet, resulting in generally higher Precision for UNet and Transformer-based Swin-UNet in Tables \ref{Table: Random Seed} and \ref{Table: Fractions}. Comparing UNet and Transformer-based Swin-UNet directly, UNet produces 0.4\% more FP than Transformer-based Swin-UNet, and Transformer-based Swin-UNet produces 0.1\% more FN than UNet, while Transformer-based Swin-UNet's TP is only 0.1\% less than UNet's. This accounts for UNet typically having a higher Recall and Transformer-based Swin-UNet a higher Precision as noted in the same tables. These differences highlight distinct trade-off strategies between precision and recall, making each model suitable for different scenarios.

The UNet model, with its high Recall, is extremely suited for wildfire monitoring scenarios in California, where missing a fire could have serious consequences. In vast and remote forest areas, high-risk meteorological regions, and near critical infrastructure, it is crucial to detect every potential fire promptly. UNet is capable of capturing as many real fire events as possible, although this may accompany a higher rate of false alarms. In these scenarios, however, the ability to quickly identify and respond to potential fires outweighs the importance of reducing false alarms. The Transformer-based model, with its high Precision, is ideally suited for wildfire monitoring scenarios in California where minimizing false alarms is crucial. For example, in ecologically sensitive areas and high-value asset protection zones, frequent false alarms could lead to unnecessary interventions and resource wastage. In these cases, Transformer-based model's high precision ensures that only high-confidence fire alarms are triggered, thus optimizing resource use and minimizing environmental or community impacts. In a larger wildfire monitoring network, UNet can serve as a high-sensitivity layer, while Transformer-based Swin-UNet can act as a high-precision initial screening layer, allowing the system to capture all potential fires promptly and reduce false alarms through Transformer-based model's precise filtering. This setup leverages the complementary strengths of both models: UNet ensures no potential fire is missed, and Transformer-based Swin-UNet ensures that only high-confidence alarms are processed further. This multilayered design not only enhances the comprehensiveness of fire detection but also optimizes the effective use of resources, making it especially suitable for large-scale wildfire monitoring networks with extensive areas and precise resource management needs.

\subsection{Efficiency of Model Architectures}
Table \ref{Table: parameters} and the bars in Figures \ref{fig: random seed comparison} and \ref{fig: fraction comparison} indicate a subtle correlation between complexity and performance. Models like UNet, despite their lower computational load and parameter count, exhibited performance on par with or surpassing more complex counterparts such as ResNet and Transformer-based Swin-UNet. This challenges the notion that higher complexity equals better performance, showing that efficiency-optimized models can achieve significant accuracy.

In order to gain a deeper understanding of the underlying reasons for these results, an extensive analysis was conducted on the architectural features of the models. It was observed that Transformer-based Swin-UNet and UNet demonstrate superior performance compared to Autoencoder and ResNet. Take the average results from Table \ref{Table: Random Seed} as an example, the AUC-PR of UNet and Transformer-based Swin-UNet are 19.89\% and 17.15\% higher than that of Autoencoder, and 38.33\% and 41.57\% higher than that of ResNet. This improvement can be partially attributed to their distinctive approaches to implementing skip connections. Unlike ResNet and Autoencoder, which only use skip connections within blocks, UNet and Transformer-based Swin-UNet employ them across the encoder and decoder, improving the integration of deep and surface-level features for detecting complex wildfire spread patterns. In UNet and Transformer-based Swin-UNet, skip connections help bridge the gap between the input data and the final prediction layer, allowing for the preservation of crucial information that might otherwise be lost in deeper layers. In contrast, Autoencoder and ResNet incorporate skip connections only within their blocks to enhance information flow. This architectural choice in UNet and Transformer-based Swin-UNet is instrumental in enhancing the models' ability to accurately predict wildfires, as it ensures that both high-level and low-level features are effectively synthesized to facilitate the prediction process. The success of UNet and Transformer-based Swin-UNet can thus be partially attributed to their architecture's inherent capacity to maintain a rich, integrated feature set across the network, highlighting the importance of such structural considerations in model design.

In comparing Transformer-based Swin-UNet and UNet, Transformer-based Swin-UNet has significantly more parameters and GFlops but only slightly better predictive performance. Transformer-based Swin-UNet's complexity arises from its self-attention mechanism, which suits high-dimensional data like images but increases computational complexity due to extensive matrix operations. Additionally, Transformer-based Swin-UNet's multi-head attention mechanism boosts parameter count and complexity, demanding more computational resources. Despite Transformer-based Swin-UNet's significant computational enhancements, its modest performance gains do not offer clear advantages in all scenarios. For example, in wildfire monitoring where quick and accurate predictions are crucial, UNet is preferable for real-time prediction and emergency measures due to its efficiency. Transformer-based Swin-UNet, however, may be more suited for post-event analysis and firefighting strategy formulation due to its higher precision and longer computation time.

\subsection{Robustness Across Initialization Variants}
As presented in Figure \ref{fig: random seed comparison}, each curve represents the AUC-PR of different models on the test dataset under the same random seed. The close clustering of points for each model (marked in four colors) suggests that the AUC-PR of each model varies little under different random seeds. Testing models with different random seeds is a common method to evaluate robustness because it helps to assess the consistency of the model's performance despite variations in the initial conditions. 

The error rates of each metric are shown in Table \ref{Table: Random Seed}. These results indicate that Autoencoder, UNet, and Transformer-based Swin-UNet demonstrate relatively lower Precision and Recall error rates, which means their performance remains more consistent under different initial conditions introduced by the random seeds. This consistency is a key indicator of robustness. In contrast, the high variability of ResNet in Recall (error rate up to 35.98\%) and Precision (error rate up to 10.62\%) suggests that this model is particularly sensitive to variations in dataset features or label distributions caused by different initializations, which could lead to unstable performance in practical applications. Therefore, while considering models for this scenario, where stability and reliability are crucial, the sensitivity of ResNet to initial conditions must be carefully evaluated.

\subsection{Influence of Data Volume on Training}
As presented in Figure \ref{fig: fraction comparison}, each curve represents the AUC-PR of different models trained with the same data volume. As the volume of data increases, both UNet and Transformer-based Swin-UNet models show a trend of improvement in performance metrics such as AUC and AUC-PR. In particular, the Transformer-based Swin-UNet shows a more significant performance increment as the amount of data increases, indicating that more training data helps the UNet and Transformer-based Swin-UNet learn and generalize better. However, the improvement in performance is not linear in all cases. Especially for the ResNet model, at certain data volumes, performance improvement reaches a plateau or shows fluctuation, possibly reflecting the model's limited adaptability to data complexity \cite{Jafar2021High-speed}. Meanwhile, the Autoencoder's performance does not show a trend of improvement. This may be due to its architecture being insufficient to capture the complex features and patterns in the data, particularly with increased data volume and complexity, since it has the lowest number of parameters \cite{Alzubaidi2021Review}.

In essence, this section evaluates model performances under varying conditions, demonstrating that UNet and Transformer-based Swin-UNet outperform ResNet and Autoencoder in key metrics. Transformer-based Swin-UNet, in particular, stands out for its balanced accuracy and false alarm rate. The analysis also reveals the models' robustness and the effects of data volume. While increased data generally benefits learning, ResNet and Autoencoder do not follow this trend, highlighting the importance of model selection based on specific task requirements and data characteristics. 

\section{Interpretability Analysis} \label{xai}

\subsection{SHAP}
\begin{figure}[htbp]
    \centering
    \includegraphics[width=\textwidth]{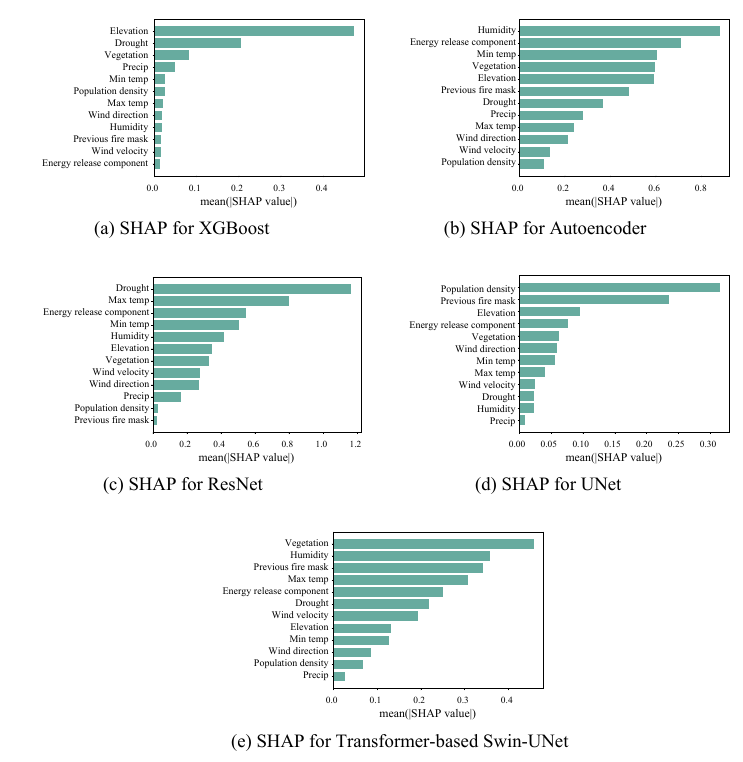}
    \caption{\modified{Mean absolute SHAP value (The average impact on model output magnitude, calculated as the mean of the absolute SHAP values for each feature across the entire dataset)}}
    \label{fig:SHAP}
\end{figure}

From the experiments conducted by Huot \cite{huot2021next}, it was discovered that removing the PFM from the training data resulted in the poorest performance of the model, thereby triggering our interest in the PFM. A clear pattern is that the degree to which models prioritize the PFM feature is closely linked to their performance in predicting wildfire spread. In the UNet and Transformer-based Swin-UNet, where PFM ranks high (second in Fig.\ref{fig:SHAP}d and third in Fig.\ref{fig:SHAP}e, respectively), we observe a correlation with these models' superior performance (AUC-PR values of 0.2739 and 0.2803, respectively). This underscores the ability of UNet and Transformer-based Swin-UNet to recognize and utilize this spatial feature, which is directly related to wildfire spread. The Autoencoder demonstrates a moderate level of performance (AUC-PR = 0.2338), with PFM's moderate ranking in Fig.\ref{fig:SHAP}b. Conversely, in the XGBoost and ResNet models, where PFM ranks low (third from last in Fig.\ref{fig:SHAP}a and last in Fig.\ref{fig:SHAP}c, respectively), there is an alignment with their lower prediction performance (AUC-PR values of 0.06 and 0.1980), suggesting these models may not fully leverage the PFM feature. This further affirms the viewpoint that PFM is a crucial feature for wildfire spread prediction.

Except for PFM, we can also find other points worth analyzing from Figure \ref{fig:SHAP}. UNet ranked 'Population density' higher than other models (1st place in Fig.\ref{fig:SHAP}d), which indicates its strong reliance on this feature for prediction. Notably, despite 'Population density''s low importance in Transformer-based Swin-UNet (11th place in Fig.\ref{fig:SHAP}e), its AUC-PR score is slightly higher than that of UNet. Transformer-based Swin-UNet prioritizes 'Vegetation' and 'Humidity' as the most important features, while UNet assigns them medium level of importance. This distinction may explain the slight advantage of Transformer-based Swin-UNet over UNet in wildfire spread prediction tasks. Transformer-based Swin-UNet's self-attention mechanism allows for a deep understanding of complex relationships between features and spatial context, especially the impact of 'Vegetation' on the spread of fire. This capability may enable Transformer-based Swin-UNet to outperform in capturing these critical dynamics. UNet outperforms in image segmentation and capturing spatial data, yet Transformer-based Swin-UNet surpasses it by integrating climate factors more flexibly through its self-attention mechanism. Aside from the previously discussed features, the remaining ones are considered variably across models. Their impact on wildfire prediction is nuanced, with each contributing to understanding the complex interplay of factors influencing fire spread. However, their relative importance and the way they are integrated vary, reflecting the diverse approaches of models to balance spatial, climatic, and human factors in predicting wildfire dynamics.

\subsection{Grad-CAM}

Following the macroscopic SHAP analysis, we identified several intriguing findings that warrant further investigation. For example, in some models, the importance of 'Population density' was notably high, while others exhibited an imbalance in attention to crucial features. To investigate deeper, we employed IG and Grad-CAM for a more detailed analysis of individual samples.From Table \ref{ig-grad-5} to Table \ref{ig-grad-125}, we present a comprehensive comparison of IG feature contribution and Grad-CAM attention heatmaps, along with visualizations of the corresponding features. By integrating these analytical tools, we hope to gain a thorough understanding of the feature contributions and model behavior specific to this dataset. 

However, before taking IG analysis into consideration, we can already find some patterns from SEG-Grad-CAM: Firstly, from Table \ref{ig-grad-5}, \ref{ig-grad-15}, \ref{ig-grad-29}, and \ref{ig-grad-125}, the Autoencoder's lack of attention towards the PFM is evidenced by predominantly blue regions, indicating a deficiency in capturing PFM features. In contrast, UNet and Transformer-based Swin-UNet exhibit a heightened focus on PFM, as shown by yellow or red areas, demonstrating their superior ability to identify and emphasize PFM. \modified{This difference underscores the capability of UNet and Transformer-based Swin-UNet to effectively recognize and prioritize PFM features, which are critical for model performance.} Secondly, UNet and Transformer-based Swin-UNet's ability to concentrate on PFM does not detract from their attention to other important features such as 'Vegetation' and 'Drought'. This is evident from the detailed contours for these features in the 'Combined Grad-CAM Heatmap' of Table \ref{ig-grad-5}, \ref{ig-grad-15}, \ref{ig-grad-17}, \ref{ig-grad-21}, \ref{ig-grad-29}, which are more pronounced than those observed in the Autoencoder and ResNet models. However, in most cases, Transformer-based Swin-UNet is more effective than UNet at capturing and retaining information on variables such as 'Vegetation' and 'Drought'. \modified{Although ResNet can also capture these features to some extent, its performance is hindered by an overemphasis on PFM, which may lead to an imbalance in capturing other critical climate variables.} Transformer-based Swin-UNet's superiority may be due to its integration of a global attention mechanism and skip connections in its advanced architecture. The global attention mechanism allows Transformer-based Swin-UNet to dynamically identify and focus on the most important information throughout the entire image, which is particularly beneficial for processing variables with complex spatial and temporal distribution features, such as 'Vegetation' and 'Drought'.

\begin{figure}[ht!]
    \centering
    \begin{minipage}{0.49\textwidth}
        \centering
        \includegraphics[width=\textwidth]{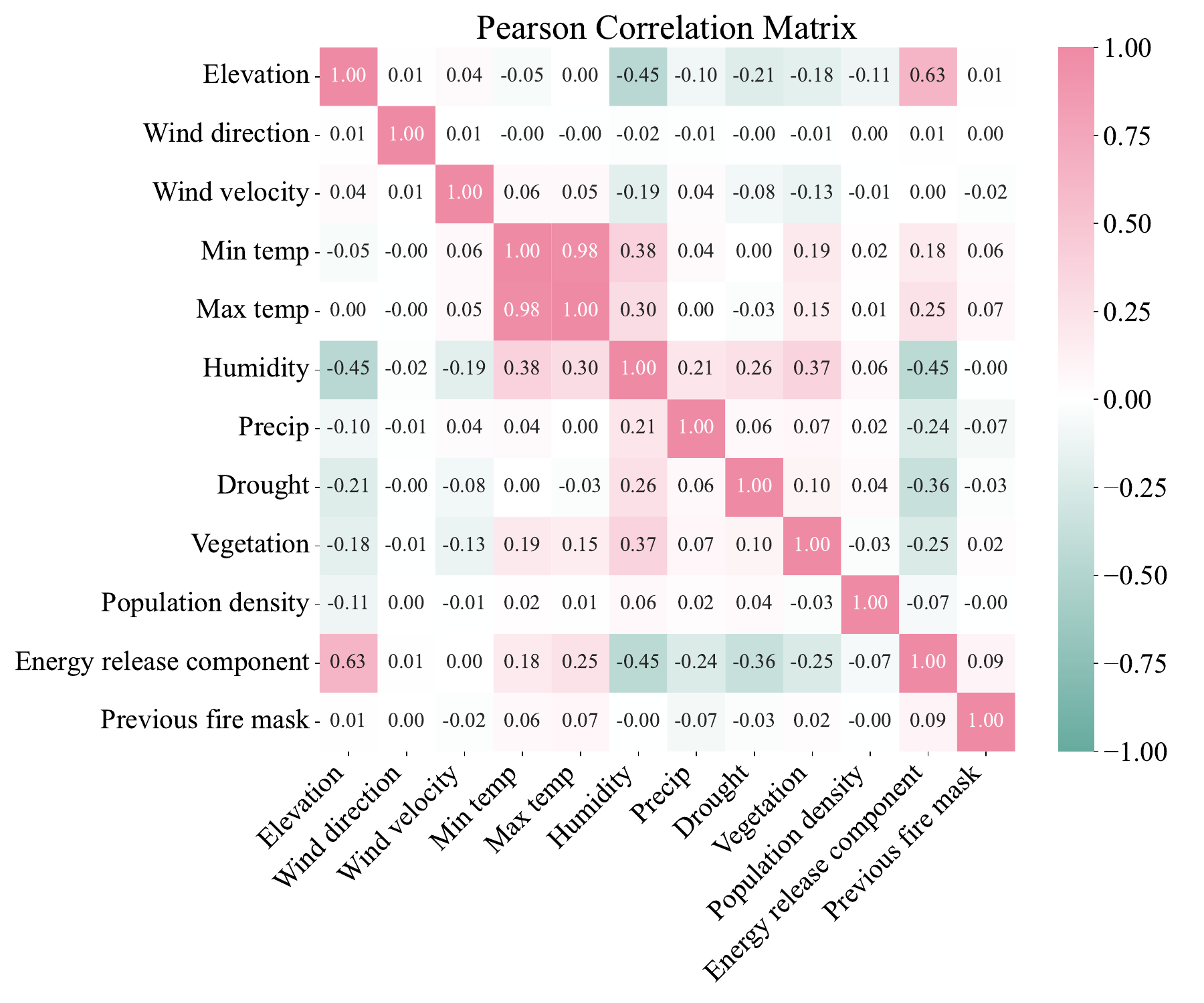}
        \caption{This matrix visualizes the pairwise Pearson correlation coefficients between features. Shades of pink represent positive correlations approaching 1, while shades of green indicate negative correlations nearing -1. The color gradient provides a clear visual indication of the strength and direction of correlations.}
        \label{fig: pearson}
    \end{minipage}\hfill
    \begin{minipage}{0.49\textwidth}
        \centering
        \includegraphics[width=\textwidth]{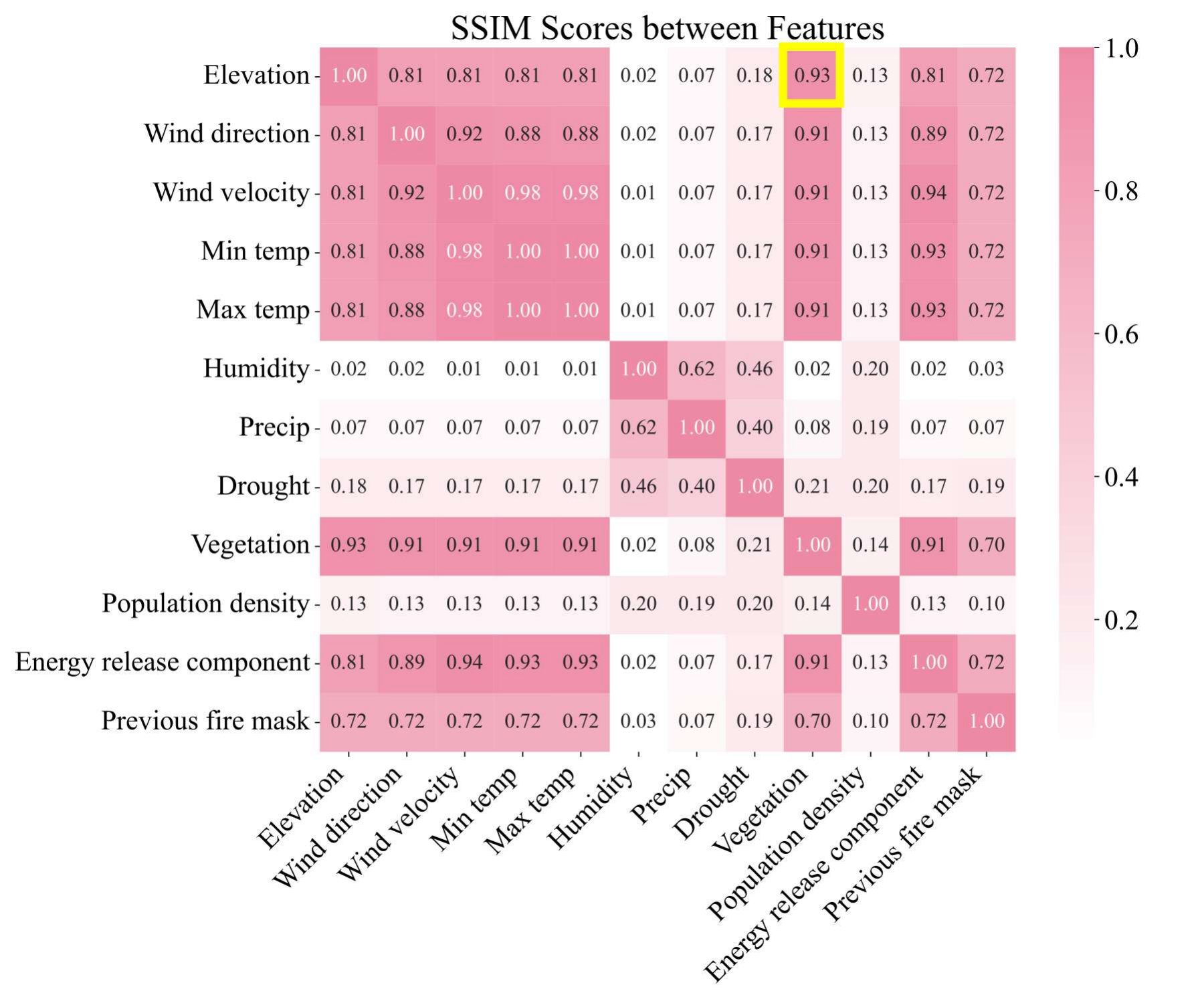}
        \caption{SSIM scores between features. This matrix displays the SSIM scores, which measure the similarity between different features. The score for 'Elevation' against 'Vegetation' is highlighted in a yellow box. Shades of pink indicate a score of 1, representing perfect similarity, while \remove{shades of green} \changed{white} represent a score of 0, indicating no similarity.}
        \label{fig: ssim}
    \end{minipage}
\end{figure}

Additionally, skip connections ensure the effective transfer and preservation of important detail information between the deep layers of the model, which helps in more accurately reconstructing these critical climate data features during the decoding phase. Therefore, Transformer-based Swin-UNet's performance in capturing and retaining information on variables like 'Vegetation' and 'Drought' seems surpass that of Autoencoder, ResNet, and UNet. Furthermore, the analysis highlights that models shift their attention towards the 'Population density' when PFM is missing. In such cases, as observed in Table \ref{ig-grad-17}, \ref{ig-grad-24}, \ref{ig-grad-25}, and \ref{ig-grad-33}, the 'Combined Grad-CAM Heatmap' of all four models shows contours of the 'Population density', which typically does not appear in cases where the PFM is present normally. Lastly, 'Elevation' is typically an important feature, but it is difficult to discern its outline in the 'Combined Grad-CAM Heatmap'. By analyzing the features in Tables \ref{ig-grad-15}, \ref{ig-grad-17}, \ref{ig-grad-21}, \ref{ig-grad-24}, \ref{ig-grad-25}, \ref{ig-grad-29}, \ref{ig-grad-33}, and \ref{ig-grad-125}, we find that 'Elevation' often shows a high similarity with the feature 'Vegetation', which is also evidenced by the SSIM evaluation. As illustrated in Figure \ref{fig: ssim}, the SSIM score of 'Elevation' against 'Vegetation', highlighted in a yellow box, is 0.93, indicating a high visual similarity between the two features. Thus, we speculate that this high degree of similarity might lead to difficulties in distinguishing between these two features in the heatmap generation. This could explain why 'Elevation' is not prominently visible in the heatmap, as the model may blend its information with that of 'Vegetation', thereby masking its unique visual identity. A similar analysis could be obtained by the pearson correlation coefficients as shown in figure \ref{fig: pearson}.

\noindent
\begin{table}
    \centering
    \caption{Analysis of the 5th sample in the dataset$^{a}$}
    \makebox[\textwidth][c]{ 
        \resizebox{1.3\textwidth}{!}{ 
            \begin{tabular}{c} 
                \includegraphics[width=\textwidth]{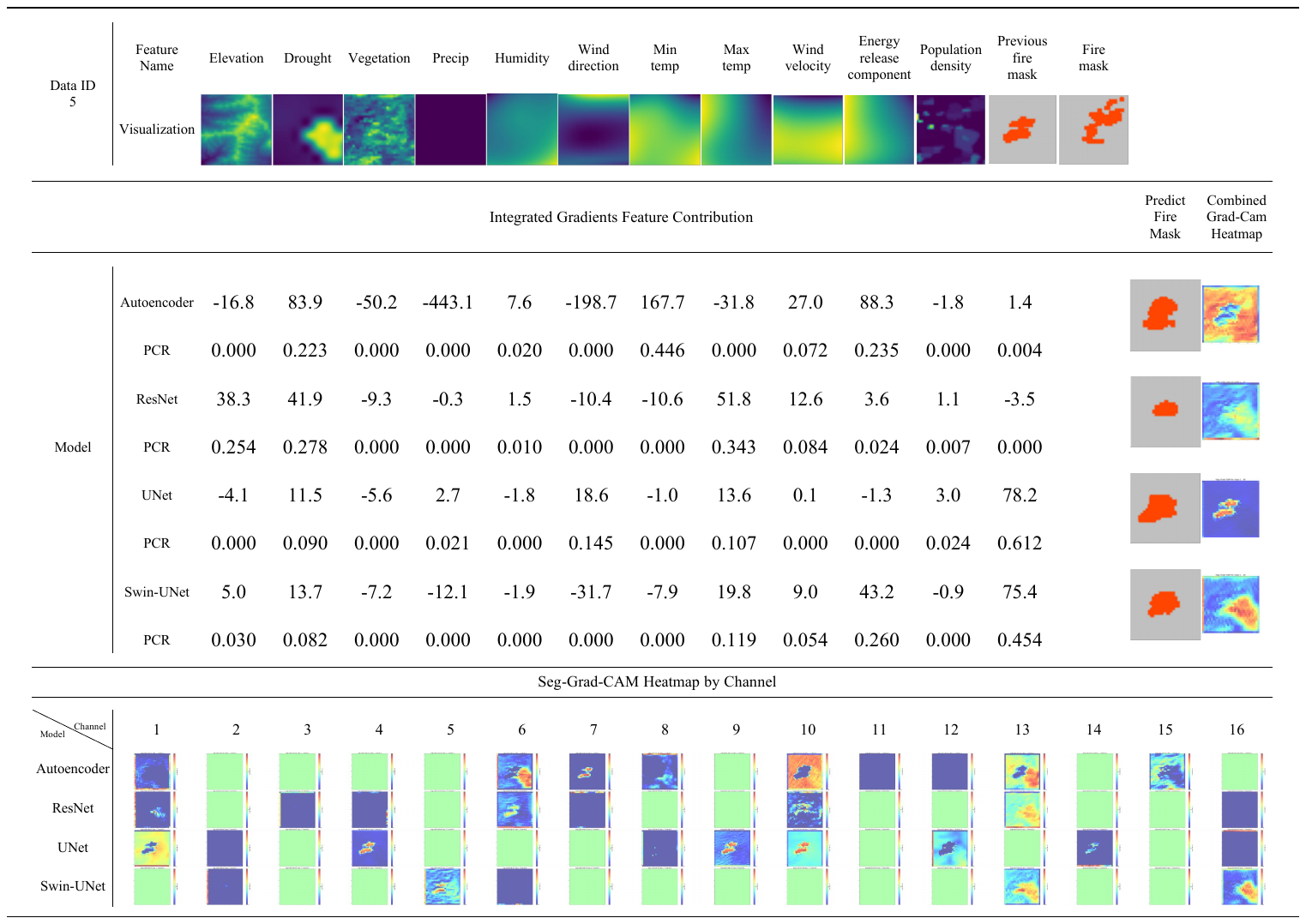} \\
                \parbox{0.8\textwidth}{
                    \footnotesize $^{a}$\remove{In the SEG-Grad-CAM attention heatmap by channel analysis} \changed{In the analysis of SEG-Grad-CAM attention heatmaps by channel (16 subfigures representing the 16 channels of the first convolutional layer for each model), the heatmaps illustrate the recognition of feature patterns by each individual channel.} It is evident that models such as Autoencoder, ResNet, UNet, and Transformer-based Swin-UNet can capture PFM, 'Drought', and 'Vegetation' to a certain extent. However, in the combined attention heatmap, the Autoencoder and ResNet models exhibit a lack of the PFM feature, which could be attributed to the models' attention being diverted by other features. In IG analysis, the feature contribution of PFM is significantly higher in UNet and Transformer-based Swin-UNet, in contrast to Autoencoder and ResNet, where it is considerably lower. Less critical features, such as 'Drought', disproportionately capture attention in the Autoencoder and ResNet models.
                }
                \label{ig-grad-5}
                
            \end{tabular}
        }
    }
\end{table}
\noindent
\begin{table}
    \centering
    \caption{Analysis of the 15th sample in the dataset$^{b}$}
    \makebox[\textwidth][c]{ 
        \resizebox{1.3\textwidth}{!}{ 
            \begin{tabular}{c} 
                \includegraphics[width=\textwidth]{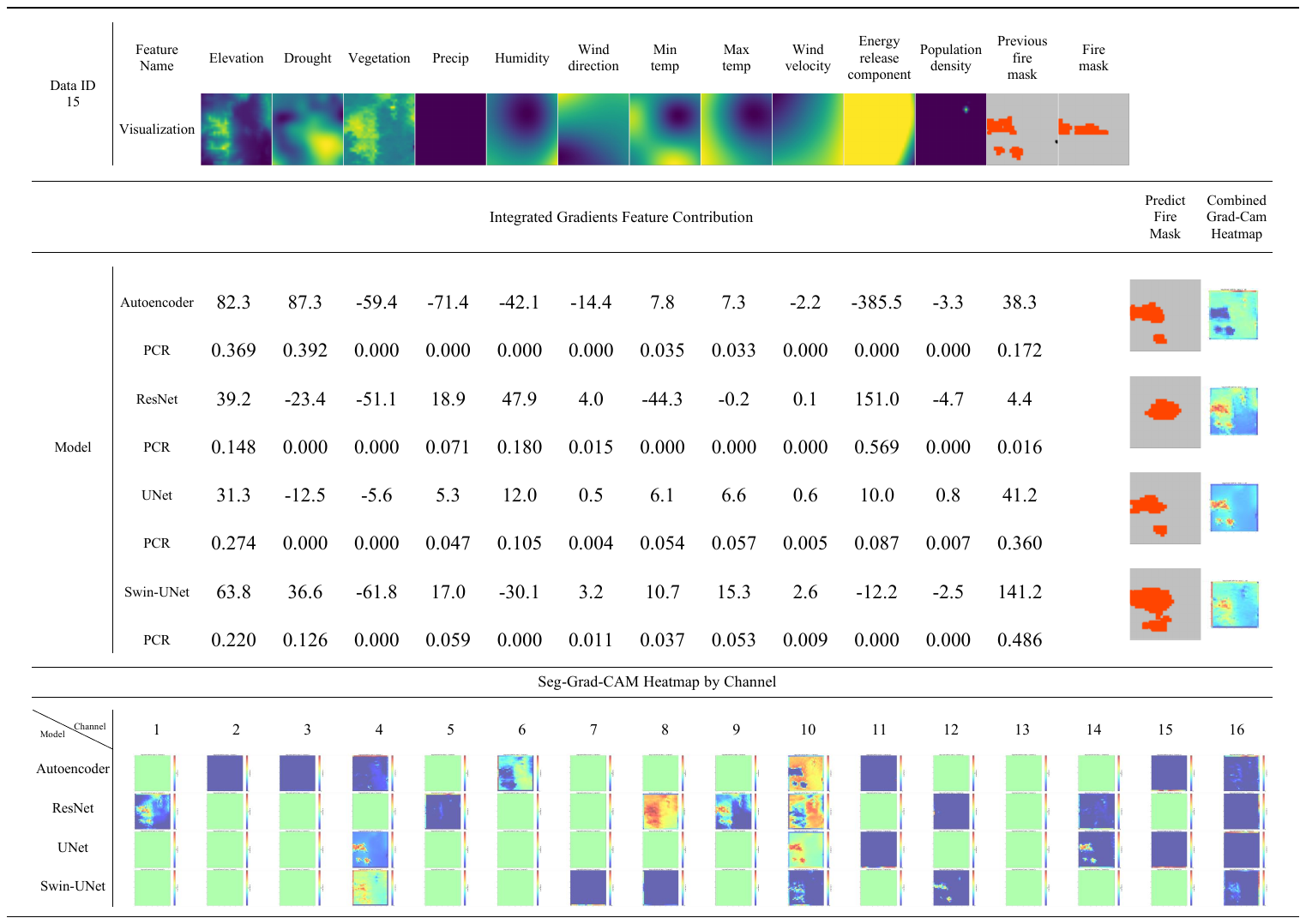} \\
                \parbox{0.8\textwidth}{
                    \footnotesize $^{b}$\remove{In the analysis of SEG-Grad-CAM attention heatmaps by channel,}\changed{In the analysis of SEG-Grad-CAM attention heatmaps by channel (16 subfigures representing the 16 channels of the first convolutional layer for each model), the heatmaps illustrate the recognition of feature patterns by each individual channel.} It is observed that Autoencoder barely captures the features of the PFM, while ResNet, UNet, and Transformer-based Swin-UNet manage to detect PFM, 'Drought', and 'Vegetation' to a certain extent. In the combined attention heatmap, the absence of PFM features in Autoencoder suggests that the model's focus may have been diverted by other features. In contrast, ResNet, UNet, and Transformer-based Swin-UNet effectively highlight the PFM along with features such as 'Vegetation'. IG analysis reveals a high feature contribution for PFM in UNet and Transformer-based Swin-UNet, whereas it is notably low in Autoencoder and ResNet. Less critical features like 'Drought' disproportionately occupy the attention of Autoencoder and ResNet.
                }
                \label{ig-grad-15}
                
            \end{tabular}
        }
    }
\end{table}

\modified{Grad-CAM provides an intuitive visual understanding for features with easily recognizable shapes, such as 'Elevation', 'Drought', 'Vegetation', and 'Population density'. However, it is less effective for features that share similarities and abstract features like 'Humidity' and 'Min temp' . These limitations arise because similar features tend to blend together and abstract features lack distinct shapes, making them difficult to distinguish on heatmaps.} Therefore, we used IG to help analyze these abstract features and quantitatively assess the contribution of each feature.

\subsection{Integrated Gradients}
\modified{Through IG analysis, we can not only better quantify abstract features that are difficult to distinguish in Grad-CAM heatmaps but also cross-validate the analyses from Grad-CAM and SHAP.} In UNet and Transformer-based Swin-UNet, the PCR of the PFM feature is typically greater compared to other models (see Table \ref{ig-grad-5}, \ref{ig-grad-15}, \ref{ig-grad-17}, \ref{ig-grad-21}, \ref{ig-grad-29}, and \ref{ig-grad-125}). This indicates that these architectures may place more emphasis on PFM. UNet and Transformer-based Swin-UNet also focus on features like 'Drought' and 'Elevation', but usually they don't surpass the crucial PFM (see Table \ref{ig-grad-15}, \ref{ig-grad-17}, and \ref{ig-grad-125}). This suggests a balanced recognition and utilization of various features, with PFM remaining crucial in their predictive decision-making. Notably, this emphasis on PFM aligns with the findings presented by \citeA{huot2021next}, which underscore the critical role of historical fire information in predicting future wildfire spread. The convergence of insights from SHAP, IG, and Grad-CAM analyses further strengthens our understanding of feature importance. For instance, the high feature importance of the PFM in SHAP analysis for UNet is mirrored in the IG and Grad-CAM analyses, where its significance is visually and quantitatively confirmed.

In Autoencoder and ResNet, when 'Drought' or 'Vegetation' is significant, they often overshadow PFM in terms of PCR (see Table \ref{ig-grad-5}, \ref{ig-grad-15}, \ref{ig-grad-17}, \ref{ig-grad-21}, \ref{ig-grad-25}, and \ref{ig-grad-29}). This implies a diverted focus from other features, leading to less effective accommodation by the Autoencoder. On the contrary, Transformer-based Swin-UNet and UNet achieve a better balance, meaning that while focusing on other features such as 'Drought', 'Vegetation', and 'Elevation', these models still maintain sufficient attention on PFM. This point also corresponds with the findings in Grad-CAM, which highlights a crucial aspect of model tuning. 
\modified{The ability of UNet and Transformer-based Swin-UNet to maintain balanced attention to critical features contributes significantly to their superior performance. This balance ensures that the models do not overly prioritize certain features at the expense of others, facilitating a more accurate and comprehensive understanding of the factors driving the spread of wildfires. UNet and Transformer-based Swin-UNet achieve this through skip connections, which help preserve important details by combining shallow and deep features. These findings provide valuable insights for future model selection and design. Incorporating skip connections can enhance a model’s ability to retain essential features across layers. This strategy facilitates the development of models that offer a more accurate and comprehensive understanding of multifaceted environmental phenomena, such as wildfire spread.}

These congruence across interpretability tools not only validates the models' reliance on critical features but also exemplifies how SHAP, IG, and Grad-CAM can collectively enhance our interpretability framework. Their combined application enables a comprehensive and detailed exploration of feature contributions, facilitating more transparent and explainable machine learning models in environmental science.

\section{Conclusion} \label{conclusion}
This study began a comprehensive exploration of deep learning models' capabilities in predicting wildfires using remote sensing data, emphasizing the importance of model interpretability. Rather than arbitrarily selecting models, our choice was aligned with mainstream machine learning methods that predominantly employ either CNN or Transformer-based architectures. This alignment is important as these architectures represent the cutting edge in handling the complex spatial data typical of remote sensing applications. Through this, we analyzed the performances and interpretability of prominent representatives from these categories—specifically, Autoencoder, ResNet, and UNet from CNNs, and Transformer-based Swin-UNet from Transformer architectures. Our findings provide detailed insights into their predictive performances and the critical role of interpretability in their application to wildfire prediction.

Firstly, our findings indicate that the UNet and Transformer-based model exhibit superior predictive accuracy compared to the Autoencoder and ResNet models. This superior performance can be attributed, in part, to their different implementations of skip connections, which facilitate effective feature transmission across network layers, enhancing the model's ability to learn and generalize from complex spatial data, and maintain focus on critical features such as PFM. Secondly, in the comparison between Transformer-based Swin-UNet and UNet, Transformer-based Swin-UNet slightly outperforms due to its adoption of a global attention mechanism and multi-head attention mechanism. These mechanisms enable Transformer-based Swin-UNet to more comprehensively capture and analyze high-dimensional data, such as climate features. This explains why Transformer-based Swin-UNet performs better than UNet in certain scenarios, as it can integrate and interpret large-scale spatial data better. Lastly, the application of interpretability tools such as SHAP, IG, and Grad-CAM has provided deeper insights into the decision-making processes of these models. These tools highlight the importance of the PFM feature and the need for balanced feature representation to enhance prediction accuracy. Besides, Grad-CAM analysis shows that Transformer-based Swin-UNet's attention mechanism not only focuses on crucial features like PFM but also effectively captures other significant variables such as 'Drought' and 'Vegetation'. This capability allows Transformer-based Swin-UNet to more accurately predict wildfire spread and development, demonstrating its unique strengths in feature balance and information extraction.

Through our comparison of models, we also provide guidance for model selection to accommodate different situations. Firstly, UNet and Transformer-based Swin-UNet offer distinctive trade-offs between precision and recall, making each suitable for specific scenarios. \modified{One reason for the difference is the implementation of the self-attention mechanism in Transformer-based Swin-UNet. The self-attention mechanism enables Swin UNet to more flexibly capture global dependencies when processing input data, helping the model to more accurately distinguish important features. This contributes to its superior performance in precision, thereby making Transformer-based Swin-UNet optimal for reducing false alarms in sensitive areas and avoiding unnecessary interventions. UNet, with its high recall, is ideal for critical areas where missing a fire could be catastrophic, despite a higher false alarm rate.} Secondly, the complexity of the UNet and Transformer-based Swin-UNet models also makes them suitable for different scenarios. Transformer-based Swin-UNet has significantly more parameters and GFlops but only slightly better predictive performance, making UNet preferable for real-time prediction and emergency measures due to its efficiency in wildfire monitoring where quick and accurate predictions are crucial. Transformer-based Swin-UNet, however, may be more suited for post-event analysis and firefighting strategy formulation due to its higher precision and longer computation time. Lastly, from our robustness experiments with the four models, we believe ResNet, due to its lack of robustness, should be carefully considered for this scenario, as stability and reliability are essential.

Future research could investigate refining model architectures to enhance both predictive accuracy and interpretability. Investigating hybrid models that take the strengths of UNet and Transformer-based models together may offer more robust predictive capabilities for wildfire forecasting. Further scrutiny of the role of skip connections in processing this particular dataset, alongside trials with other models incorporating such features, is warranted. Moreover, the slight advantage of Transformer-based Swin-UNet over UNet—whether it stems from the transformer structure's advantages or merely a greater number of parameters—merits further investigation. Additionally, enhancing the use of interpretability tools in model assessment could promote increased trust and transparency in deploying deep learning for environmental surveillance and disaster response.

\newpage
\appendix
\section{Tables for IG and Grad-CAM Analysis}
\noindent
\begin{table}[H]
    \centering
    \caption{Analysis of the 17th sample in the dataset$^{c}$}
    \makebox[\textwidth][c]{ 
        \resizebox{1.2\textwidth}{!}{ 
            \begin{tabular}{c} 
                \includegraphics[width=\textwidth]{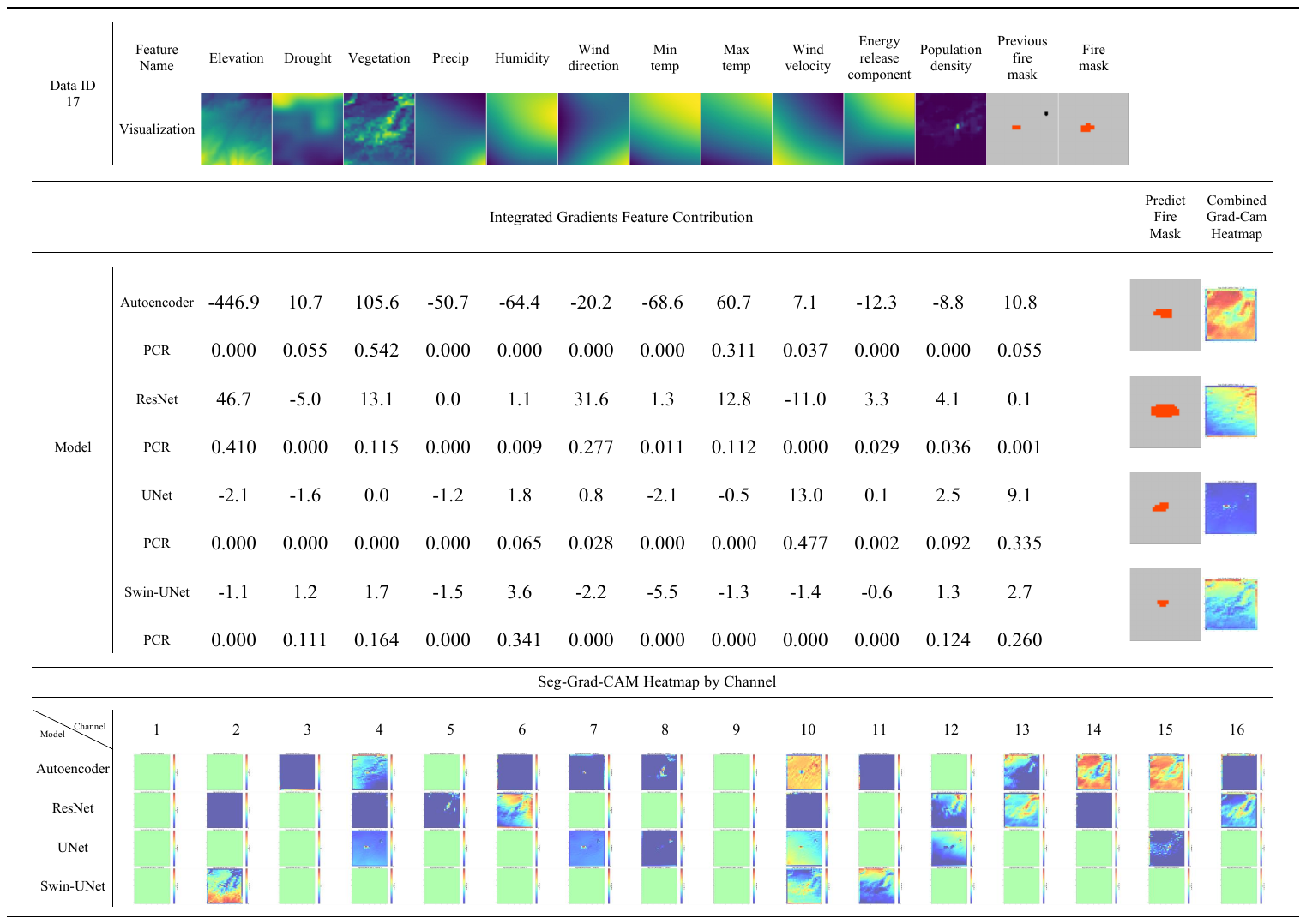} \\
                \parbox{0.8\textwidth}{ 
                    \footnotesize $^{c}$In the analysis of SEG-Grad-CAM attention heatmaps by channel, it appears that possibly due to the small size of the PFM, Autoencoder, ResNet, and Transformer-based Swin-UNet have almost failed to capture its features, with UNet being the exception. It is evident that all four models have successfully identified features related to 'Drought' and 'Vegetation'. In the combined attention heatmap, UNet clearly captures the PFM, while the other three models barely show the PFM shape, likely due to its small size. IG analysis indicates that the feature contribution of PFM remains significantly high in UNet and Transformer-based Swin-UNet, but notably low in Autoencoder and ResNet. Less critical features, such as 'Vegetation', have disproportionately drawn the attention of Autoencoder and ResNet.
                }
                \label{ig-grad-17}
            \end{tabular}
        }
    }
\end{table}
\noindent
\begin{table}[H]
    \centering
    \caption{Analysis of the 21th sample in the dataset$^{d}$}
    \makebox[\textwidth][c]{ 
        \resizebox{1.2\textwidth}{!}{ 
            \begin{tabular}{c} 
                \includegraphics[width=\textwidth]{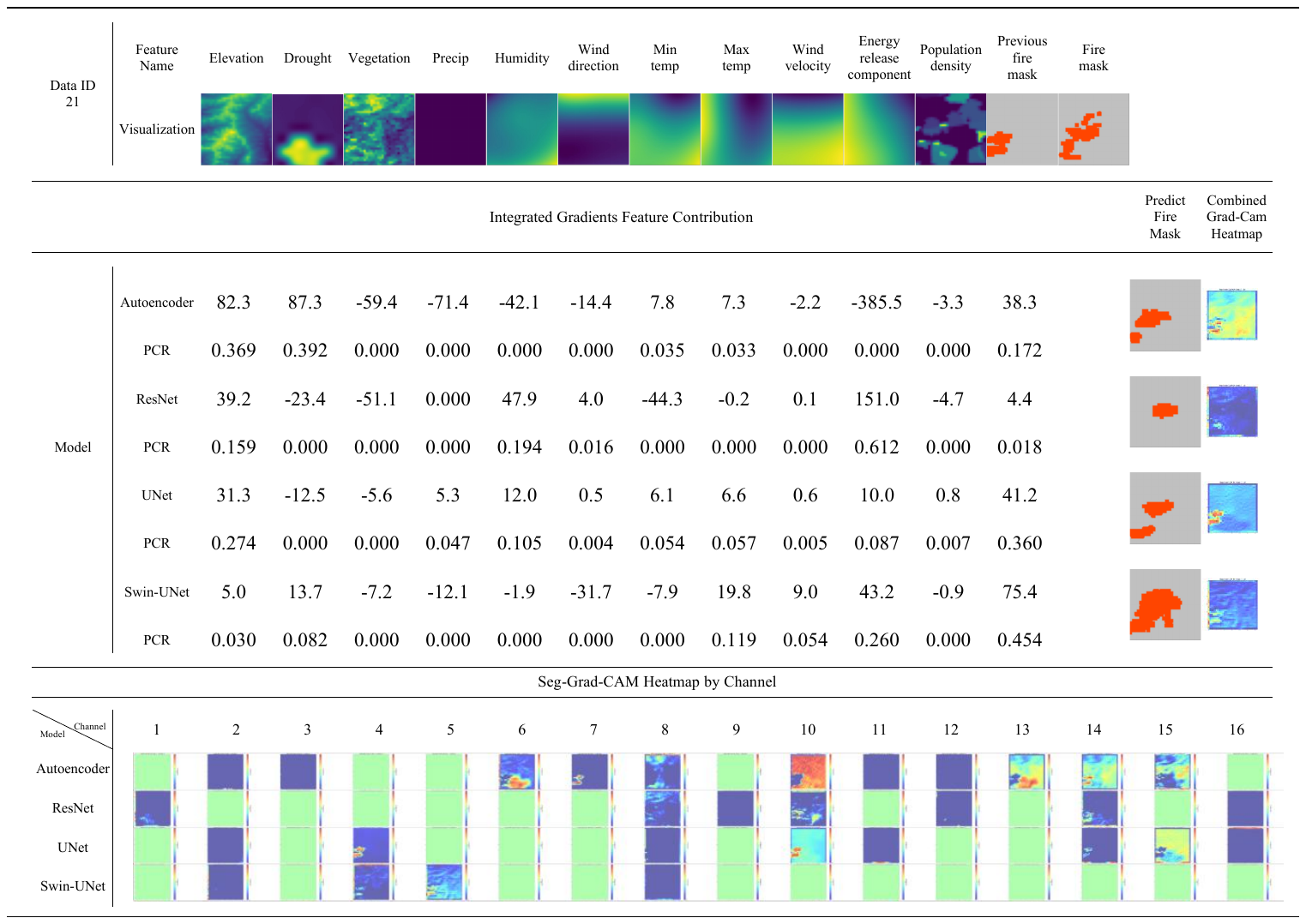} \\
                \parbox{0.8\textwidth}{ 
                    \footnotesize $^{d}$In the SEG-Grad-CAM attention heatmaps by channel, it is observed that Autoencoder, ResNet, UNet, and Transformer-based Swin-UNet are capable of capturing features like the PFM, 'Drought', and 'Vegetation' to varying degrees. In the combined attention heatmap, Autoencoder shows a notable absence of PFM features, and ResNet displays these features faintly, which may be due to the models' attention being diverted by other features. Conversely, UNet and Transformer-based Swin-UNet outperform in highlighting both PFM and 'Vegetation' features prominently. IG analysis further reveals that the feature contribution of PFM is significantly higher in UNet and Transformer-based Swin-UNet, whereas it remains quite low in Autoencoder and ResNet. Less critical features such as 'Drought' occupy a substantial amount of attention in Autoencoder and ResNet models.
                }
                \label{ig-grad-21}
            \end{tabular}
        }
    }
\end{table}
\noindent
\begin{table}
    \centering
    \caption{Analysis of the 24th sample in the dataset$^{e}$}
    \makebox[\textwidth][c]{ 
        \resizebox{1.3\textwidth}{!}{ 
            \begin{tabular}{c} 
                \includegraphics[width=\textwidth]{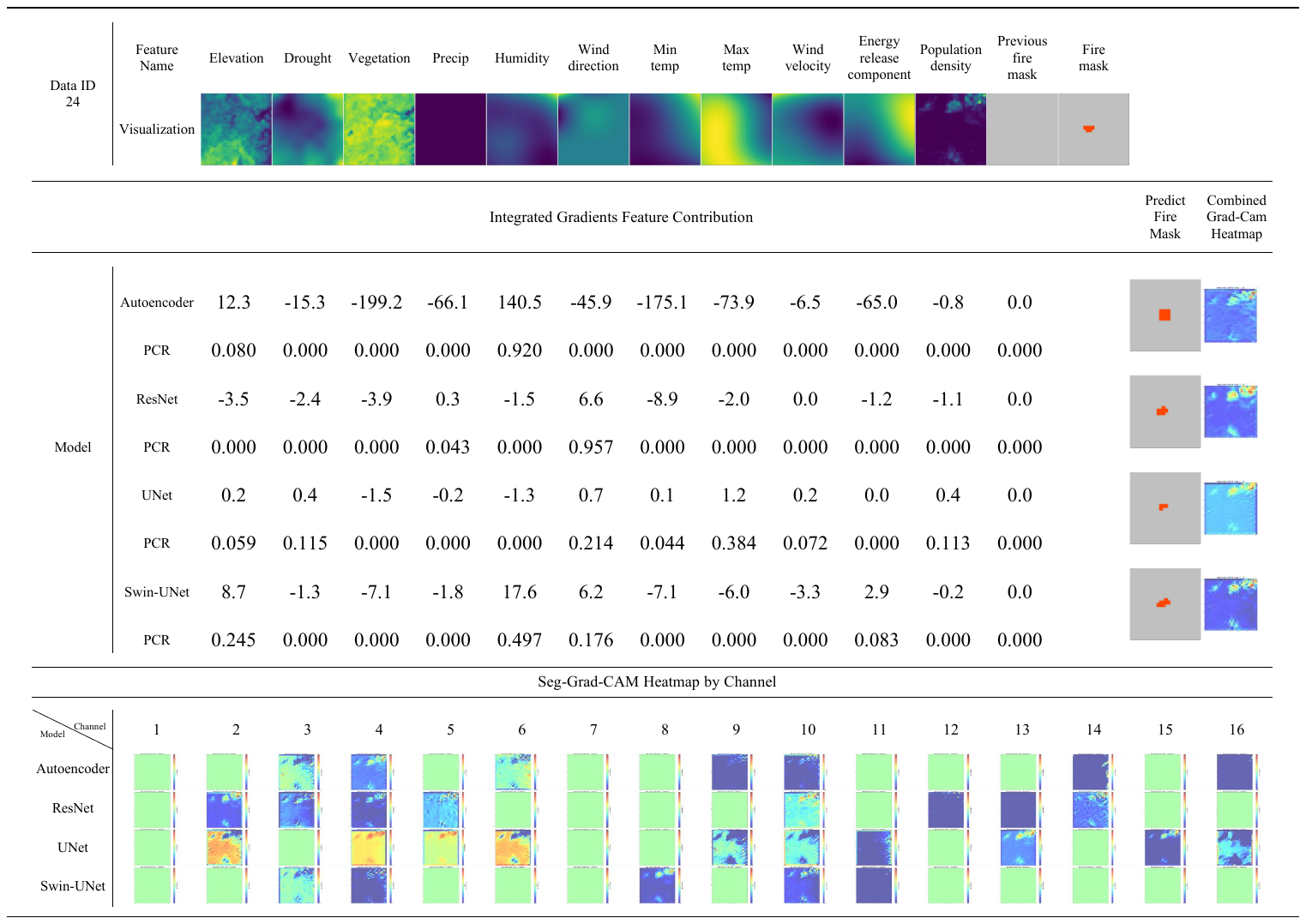} \\
                \parbox{0.8\textwidth}{ 
                    \footnotesize $^{e}$Analysis of the 24th data. In the analysis of SEG-Grad-CAM attention heatmaps by channel, the absence of PFM is attributed to the lack of PFM data. It is observed that Autoencoder, ResNet, UNet, and Transformer-based Swin-UNet are all capable of detecting features like 'Drought' and 'Vegetation' to some extent. Notably, 'Population density' feature, which was absent in most of the samples, is now clear. In the combined attention heatmap, the primary feature captured by all four models is the 'Population density'. According to the IG analysis, the feature contribution of PFM is zero across all models, which is consistent with the absence of PFM data in the dataset.
                }
                \label{ig-grad-24}
            \end{tabular}
        }
    }
\end{table}
\noindent
\begin{table}
    \centering
    \caption{Analysis of the 25th sample in the dataset$^{f}$}
    \makebox[\textwidth][c]{ 
        \resizebox{1.3\textwidth}{!}{ 
            \begin{tabular}{c} 
                \includegraphics[width=\textwidth]{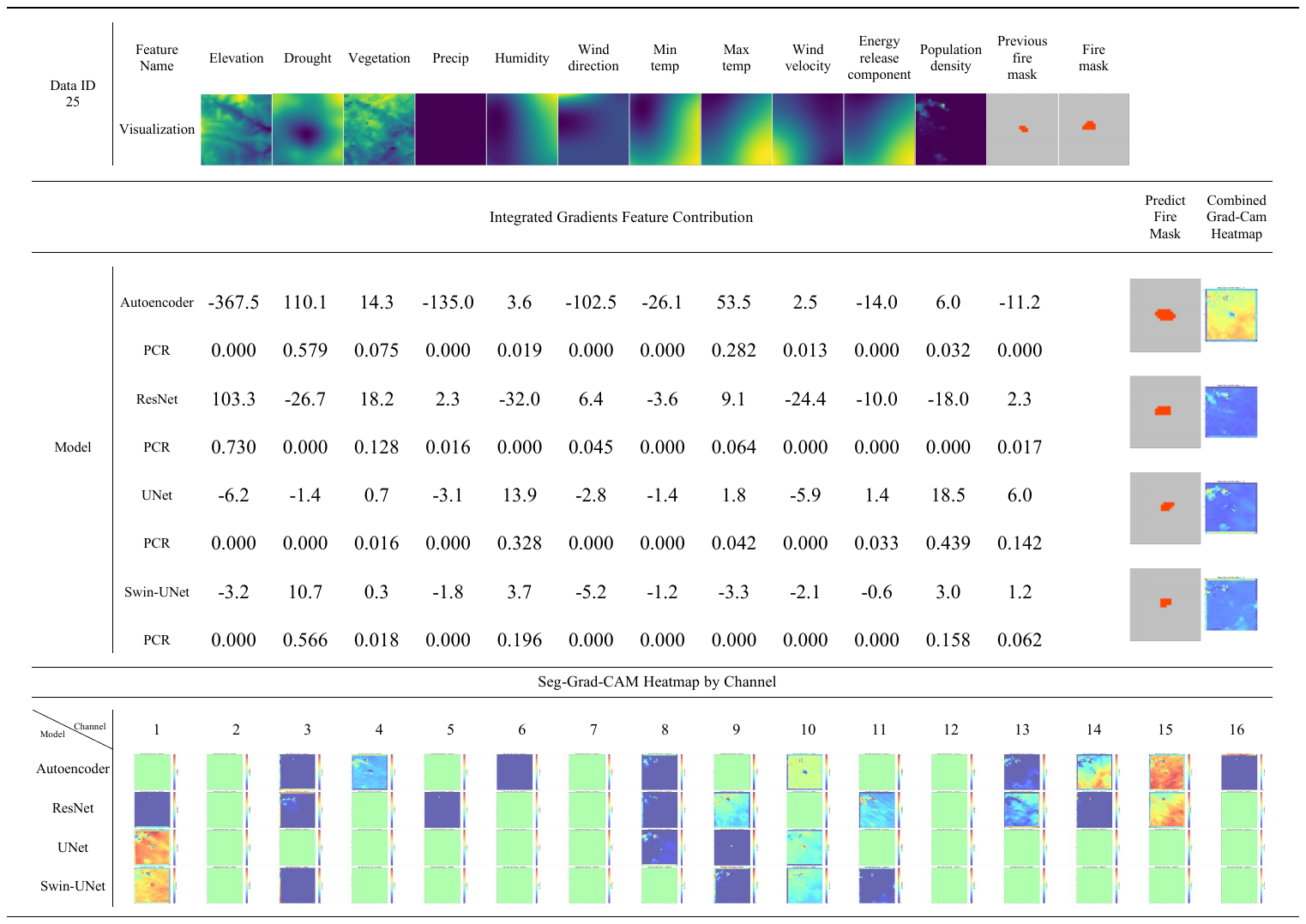} \\
                \parbox{0.8\textwidth}{ 
                    \footnotesize $^{f}$In the analysis of SEG-Grad-CAM attention heatmaps by channel, it is observed that perhaps due to the small size of the PFM, Autoencoder and ResNet almost fail to capture its features, while UNet and Transformer-based Swin-UNet can capture it. All four models successfully captured 'Drought' and 'Vegetation'. In the combined attention heatmap, UNet and Transformer-based Swin-UNet exhibit some shape of PFM, but the other two models, especially Autoencoder, show a clear absence of these features, probably attributed to the small size of PFM. IG analysis indicates that the feature contribution of PFM is not high in UNet and Transformer-based Swin-UNet, and remains low in Autoencoder and ResNet as well.
                }
                \label{ig-grad-25}
            \end{tabular}
        }
    }
\end{table}
\noindent
\begin{table}
    \centering
    \caption{Analysis of the 29th sample in the dataset$^{g}$}
    \makebox[\textwidth][c]{ 
        \resizebox{1.3\textwidth}{!}{ 
            \begin{tabular}{c} 
                \includegraphics[width=\textwidth]{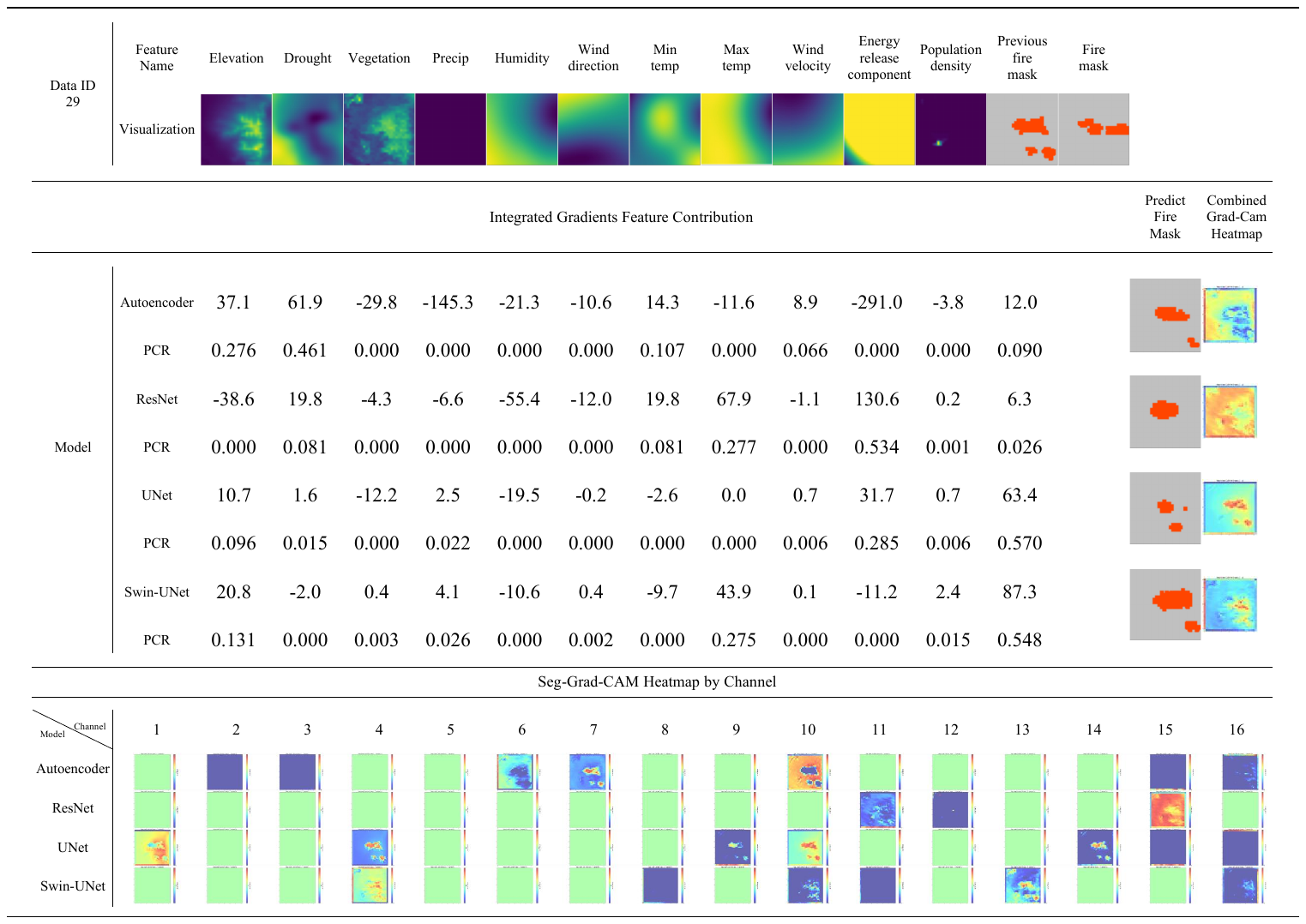} \\
                \parbox{0.8\textwidth}{ 
                    \footnotesize $^{g}$In the SEG-Grad-CAM attention heatmaps by channel, it is evident that Autoencoder, ResNet, UNet, and Transformer-based Swin-UNet are capable of capturing PFM, 'Drought', and 'Vegetation'. In the combined attention heatmap, the PFM feature is missing in Autoencoder and not clear in ResNet, possibly due to these models' focus being diverted to other features. Conversely, UNet and Transformer-based Swin-UNet effectively highlight the PFM along with 'Vegetation' features. IG analysis reveals that the feature contribution of PFM is significantly high in UNet and Transformer-based Swin-UNet, whereas it is notably low in Autoencoder and ResNet. This is in contrast to features like 'Drought', which, despite being less critical, occupy a substantial amount of attention in Autoencoder and ResNet.
                }
                \label{ig-grad-29}
            \end{tabular}
        }
    }
\end{table}
\noindent
\begin{table}
    \centering
    \caption{Analysis of the 33th sample in the dataset$^{h}$}
    \makebox[\textwidth][c]{ 
        \resizebox{1.3\textwidth}{!}{ 
            \begin{tabular}{c} 
                \includegraphics[width=\textwidth]{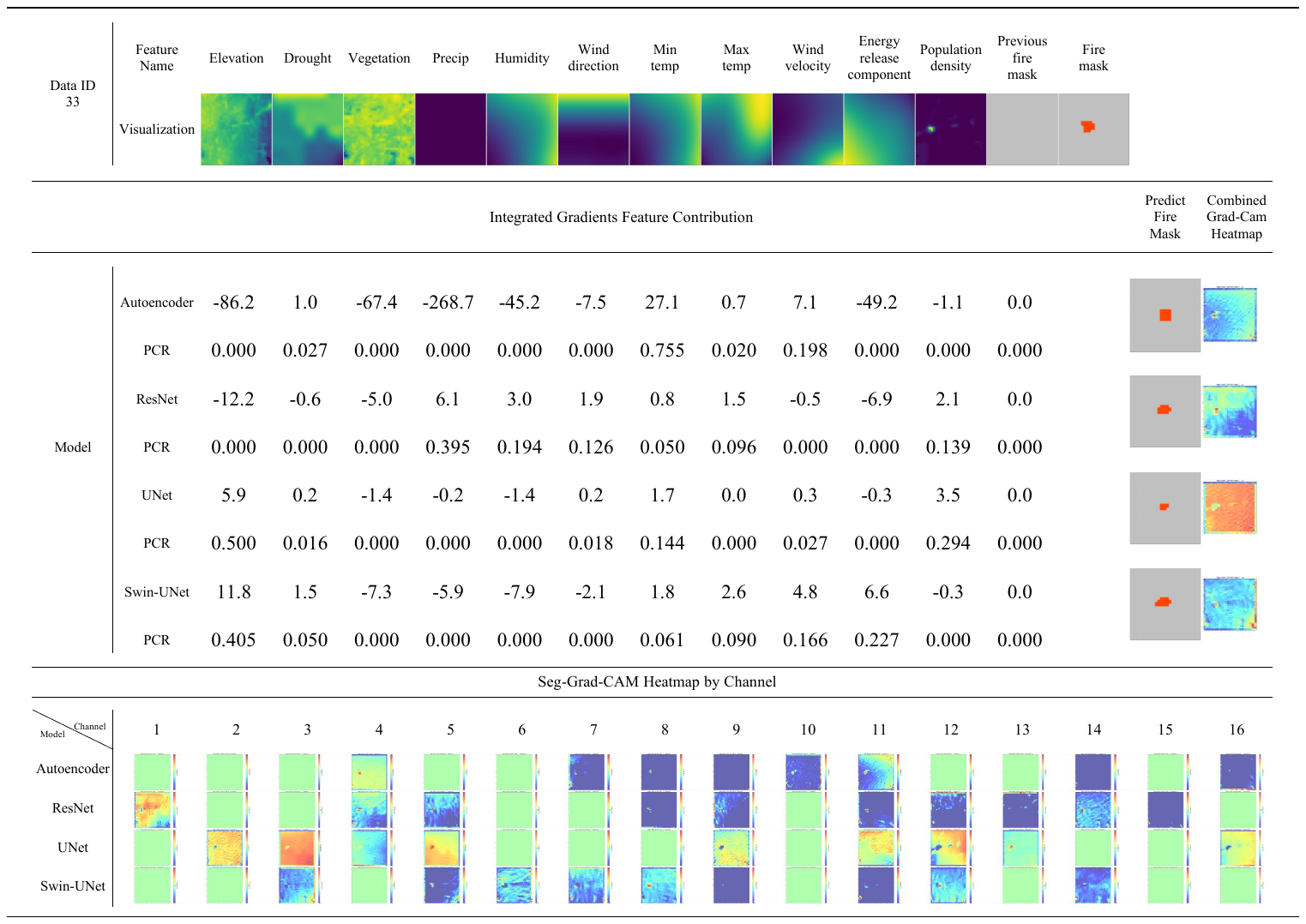} \\
                \parbox{0.8\textwidth}{ 
                    \footnotesize $^{h}$In the SEG-Grad-CAM attention heatmaps by channel, the absence of PFM is due to the lack of such data in the dataset. It is observed that Autoencoder, ResNet, UNet, and Transformer-based Swin-UNet are capable of capturing features like 'Drought' and 'Vegetation'. Notably, the 'Population density' feature, which was almost absent in other samples, is now more clear. In the combined attention heatmap, the primary feature captured by all four models is the 'Population density'. IG analysis reflects that, due to the absence of PFM data, the feature contribution of PFM is zero across all models. Meanwhile, the 'Population density' plays a more significant role in ResNet and UNet than previously observed, indicating its increased relevance in the analysis of these models.
                }
                \label{ig-grad-33}
            \end{tabular}
        }
    }
\end{table}
\noindent
\begin{table}
    \centering
    \caption{Analysis of the 125th sample in the dataset$^{i}$}
    \makebox[\textwidth][c]{ 
        \resizebox{1.3\textwidth}{!}{ 
            \begin{tabular}{c} 
                \includegraphics[width=\textwidth]{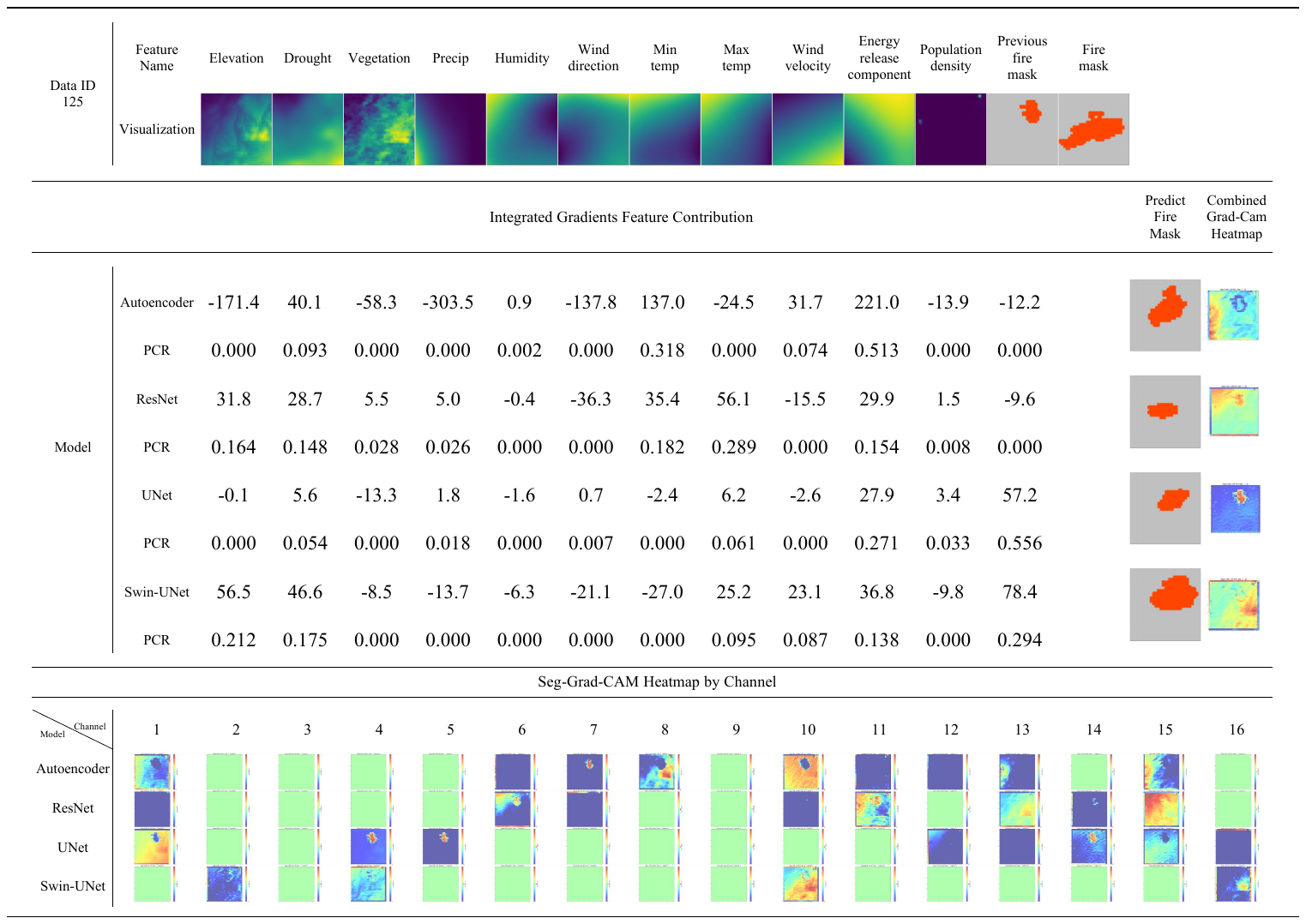} \\
                \parbox{0.8\textwidth}{ 
                    \footnotesize $^{i}$In the analysis of SEG-Grad-CAM attention heatmaps by channel, we observed that Autoencoder, ResNet, UNet, and Transformer-based Swin-UNet are capable of capturing features of PFM, 'Drought', and 'Vegetation' to varying extents. However, in the combined attention heatmap, the PFM feature is missing in Autoencoder and not clear in Transformer-based Swin-UNet, which may be attributed to these models' attention being diverted to other features. IG analysis reveals that the feature contribution of PFM is significantly high in UNet and Transformer-based Swin-UNet, whereas it is zero in Autoencoder and ResNet, indicating no positive contribution from PFM in these two models. Features such as 'Drought', which is less critical, attract more attention in Autoencoder and ResNet.
                }
                \label{ig-grad-125}
            \end{tabular}
        }
    }
\end{table}

%
%

%

%
  \begin{notation}
  \notation{\(\beta_i\)} i-th moment exponential decay rate
  \notation{\( \phi_i (v) \)} Shapley value for feature \( i \)
  \notation{\( S \)} a subset of features excluding feature \( i \)
  \notation{\( N \)} the set of all features in the model
  \notation{\( |N| \)} the total number of features
  \notation{\( |S| \)} the number of features in subset \( S \)
  \notation{\( v (S \cup \{i\}) \)} the prediction using features in \( S \) along with feature \( i \)
  \notation{\( v (S) \)} the prediction of the model using the features in subset \( S \)
  \notation{\( v (S \cup \{i\}) - v (S) \)} the marginal contribution of feature \( i \) when added to the subset \( S \)
  \notation{\( A^k \)} the \( k \)-th feature map  in a convolutional layer
  \notation{\(y^c\)} the logit corresponding to a specific class \(c\)
  \notation{\( w_k^c \)} weights associated with the \( A^k \)
  \notation{\( y_{i, j}^c\)} the logits to every pixel \( x_{i, j}\) for the class\(c\)
  \notation{\(M\)} the set of pixel indices of interest within the output mask
  \notation{\( w_{k, i, j}^c \)} pixel-specific weights associated with the \( A^k \)
  \notation{\( F\)} represents the model
  \notation{\( x_i\)} a specific pixel value within one of the 12 feature channels
  \notation{\( x'_i\)} The baseline input pixel value corresponding to \(x_i\)
  \notation{\( \lambda\)} A scaling factor ranging from 0 to 1, used to interpolate between the baseline input \(x'_i\) and the actual input \(x_i\)
  \notation{\(\gamma_i\)} the feature contribution of the \(i\)-th feature
  \notation{\(PCR_i\)} positive contribution ratio for \(i\)-th feature
  \notation{\(IG_i\)} the attribution for each input pixel value \( x_i \) against a baseline pixel value \( x'_i \)
  \end{notation}

\section*{Acknowledgements}
Sibo Cheng acknowledges the support of the French Agence Nationale de la Recherche (ANR) under reference ANR-22-CPJ2-0143-01.

\section*{Open Research}
The multivariate dataset titled `Next Day Wildfire Spread' in this study is available at Kaggle \changed{\cite{huot2022data} with the Creative Commons Attribution 4.0 International (CC BY 4.0) license.}

The codes used for implementing the model and the XAI tools for wildfire prediction and explanation, along with detailed documentation, are preserved on GitHub. \changed{These resources are accessible at \url{https://github.com/Yng314/xai_for_wildfire} under the MIT License, which permits free use, modification, and redistribution.}

\bibliography{agusample}
\end{document}